\documentclass[10pt,twocolumn,letterpaper]{article}

\PassOptionsToPackage{numbers}{natbib}
\newcommand{\citecustom}{\citep}

\usepackage{cvpr}              

\usepackage[accsupp]{axessibility}

\usepackage{graphicx}
\usepackage{amsmath}
\usepackage{amssymb}
\usepackage{booktabs}

\usepackage[utf8]{inputenc} 
\usepackage[T1]{fontenc}    
\usepackage{url}            
\usepackage{amsfonts}       
\usepackage{nicefrac}       
\usepackage{microtype}      
\usepackage{xcolor}         
\usepackage{natbib}         
\usepackage{mathtools}      %
\usepackage{comment}
\usepackage[size=scriptsize]{subcaption}
\usepackage{multirow}
\usepackage{graphbox}
\usepackage{nameref}
\usepackage{breqn}

\usepackage[pagebackref,breaklinks,colorlinks]{hyperref}

\usepackage[capitalize]{cleveref}
\crefname{section}{Sec.}{Secs.}
\Crefname{section}{Section}{Sections}
\Crefname{table}{Table}{Tables}
\crefname{table}{Tab.}{Tabs.}

\def\cvprPaperID{11751} 
\def\confName{CVPR}
\def\confYear{2023}

\begin{document}

\iffalse
    \newcommand{\cutabstractup}{\vspace*{-0.1in}}
    \newcommand{\cutabstractdown}{\vspace*{-0.1in}}

    \newcommand{\cutsectionup}{\vspace*{-0.08in}}
    \newcommand{\cutsectiondown}{\vspace*{-0.07in}}

    \newcommand{\cutsubsectionup}{\vspace*{-0.08in}}
    \newcommand{\cutsubsectiondown}{\vspace*{-0.07in}}

    \newcommand{\cutsubsubsectionup}{\vspace*{-0.08in}}
    \newcommand{\cutsubsubsectiondown}{\vspace*{-0.07in}}

    \newcommand{\cutparagraphup}{\vspace*{-0.09in}}
    \newcommand{\cutparagraphdown}{\vspace*{-0in}}

    \newcommand{\cutcaptionup}{\vspace*{-0.1in}}
    \newcommand{\cutcaptiondown}{\vspace*{-0.1in}}

    \newcommand{\cuttablecaptionup}{\vspace*{-0.1in}}
    \newcommand{\cuttablecaptiondown}{\vspace*{-0.1in}}

    \newcommand{\cutequationup}{\vspace*{-0.1in}}
    \newcommand{\cutequationdown}{\vspace*{-0.1in}}

    \newcommand{\cuttableup}{}
    \newcommand{\cuttabledown}{}

    \newcommand{\cut}{{\vspace*{-0.02in}}}
    \newcommand{\cutmore}{{\vspace*{-0.06in}}}
    \newcommand{\negcut}{}
\else 
    \newcommand{\cutsectionup}{}
    \newcommand{\cutsectiondown}{}

    \newcommand{\cutsubsectionup}{}
    \newcommand{\cutsubsectiondown}{}

    \newcommand{\cutsubsubsectionup}{}
    \newcommand{\cutsubsubsectiondown}{}

    \newcommand{\cutparagraphup}{}
    \newcommand{\cutparagraphdown}{}

    \newcommand{\cutcaptionup}{}
    \newcommand{\cutcaptiondown}{}

    \newcommand{\cutequationup}{}
    \newcommand{\cutequationdown}{}

    \newcommand{\cuttableup}{}
    \newcommand{\cuttabledown}{}

    \newcommand{\cut}{}
    \newcommand{\cutmore}{}
    \newcommand{\negcut}{}
\fi

\title{Sequential training of GANs against GAN-classifiers reveals correlated ``knowledge gaps'' present among independently trained GAN instances}

\author{%
  Arkanath Pathak \qquad Nicholas Dufour \\
  Google Research \\ 
  \texttt{\{arkanath,ndufour\}@google.com}
}

\maketitle
\begin{abstract}
Modern Generative Adversarial Networks (GANs) generate realistic images remarkably well. Previous work has demonstrated the feasibility of ``GAN-classifiers'' that are distinct from the co-trained discriminator, and operate on images generated from a frozen GAN. That such classifiers work at all affirms the existence of ``knowledge gaps'' (out-of-distribution artifacts across samples) present in GAN training. We iteratively train GAN-classifiers and train GANs that ``fool'' the classifiers (in an attempt to fill the knowledge gaps), and examine the effect on GAN training dynamics, output quality, and GAN-classifier generalization. We investigate two settings, a small DCGAN architecture trained on low dimensional images (MNIST), and StyleGAN2, a SOTA GAN architecture trained on high dimensional images (FFHQ). We find that the DCGAN is unable to effectively fool a held-out GAN-classifier without compromising the output quality. However, the StyleGAN2 can fool held-out classifiers with no change in output quality, and this effect persists over multiple rounds of GAN/classifier training which appears to reveal an ordering over optima in the generator parameter space. Finally, we study different classifier architectures and show that the architecture of the GAN-classifier has a strong influence on the set of its learned artifacts.

\end{abstract}
\section{Introduction} \label{sec:introduction}
\cutsectiondown

GAN \citecustom{goodfellow2014generative} architectures like StyleGAN2 \citecustom{karras2020analyzing} generate high-resolution images that appear largely indistinguishable from real images to the untrained eye \citecustom{lago2021more,Nightingale2022AIsynthesizedFA,hulzebosch2020detecting}. While there are many positive applications, the ability to generate large amounts of realistic images is also a source of concern given its potential application in scaled abuse and misinformation. In particular, GAN-generated human faces are widely available (e.g., \href{https://thispersondoesnotexist.com}{thispersondoesnotexist.com}) and have been used for creating fake identities on the internet \citecustom{hill2020deceive}.
 
Detection of GAN-generated images is an active research area (\emph{see} \citecustom{gragnaniello2021gan} for a survey of approaches), with some using custom methods and others using generic CNN-based classifiers. Such \emph{classifiers} are distinct from the discriminator networks that are trained alongside the generator in the archetypal GAN setup. Given the adversarial nature of the training loss for GANs, the existence of the GAN-classifiers suggest consistent generator \emph{knowledge gaps} (\emph{i.e.,} artifacts present across samples that distinguish generated images from those of the underlying distribution) left by discriminators during training. Specialized classifiers \citecustom{Wang_2020_CVPR} are able to detect images sampled from held-out GAN instances and even from held-out GAN architectures. These generalization capabilities imply that the knowledge gaps are consistent not only across samples from a GAN generator but across independent GAN generator instances.
 
 In this work we modify the GAN training loss in order to fool a GAN-classifier in addition to the co-trained discriminator, and examine the effect on training dynamics and output quality. We conduct multiple rounds of training independent pools (initialized differently) of GANs followed by GAN-classifiers, and gain new insights into the GAN optimization process. We investigate two different settings: in the first setting, we choose the low-dimensional domain of handwritten digits (MNIST \citecustom{lecun2010mnist}), using a small DCGAN \citecustom{radford2015unsupervised} architecture and a vanilla GAN-classifier architecture. For the second setting, we choose a high-dimensional domain of human faces (FFHQ \citecustom{karras2019style}) with StyleGAN2 (SG2) as a SOTA GAN architecture, and three different GAN-classifier architectures (ResNet-50 \citecustom{he2016deep}, Inception-v3 \citecustom{szegedy2016rethinking}, and MobileNetV2 \citecustom{sandler2018mobilenetv2}). Our findings in this paper are as follows:
 \begin{itemize}
     \item Samples drawn from a GAN instance exhibit a space of ``artifacts'' that are exploited by the classifiers, and this space is strongly correlated with those of other GAN generator instances. This effect is present in both the DCGAN and SG2 settings.
     \item Upon introducing the need to fool held-out classifiers, the DCGAN is unable to generate high quality outputs.
     \item In the high dimensional setting, however, SG2 generators can easily fool held-out trained classifiers, and move to a new artifact space. Strikingly, we find that the artifact space is correlated among the new population of generators as it was in the original population. This correlation appears to persist in subsequent rounds as new classifiers are introduced that are adapted to the new artifact spaces.
     \item MobileNetV2 classifier instances in the SG2 setting appear unable to learn all of the artifacts available for them to exploit. Instead, MobileNetV2 instances form clusters based on the subset of artifacts learned. We hypothesize this being an effect of classifier capacity.
     \item An SG2 generator trained to reliably fool unseen classifier instances from a given architecture is not guaranteed to fool classifiers from another architecture. Therefore, the artifacts learned by a given classifier depends strongly on the classifier's architecture.
 \end{itemize}
\cutsectionup
\section{Related Work}
\cutsectiondown

Research into detection of GAN-generated media has largely tracked the increasing prominence and output quality of GANs themselves. Several studies \citecustom{mo2018fake,cozzolino2018forensictransfer,hsu2018learning,marra2018detection,tariq2018detecting,Wang_2020_CVPR,frank2020leveraging,cozzolino2021towards,gragnaniello2021gan} focus on detection of GAN-generated images using CNNs, and their robustness to data augmentation at test time. Of particular interest to us is \citecustom{Wang_2020_CVPR}, who train a ResNet-50 classifier (pre-trained using ImageNet \citecustom{russakovsky2015imagenet}) on images generated using just one modern GAN architecture, ProGAN \citecustom{karras2018progressive}. They show that the classifier generalizes to unseen GAN architectures, concluding that the task of general GAN detection is fairly straightforward, at least in the absence of image augmentations. Later studies dispute this \citecustom{gragnaniello2021gan,frank2021re}, demonstrating that the test performance of the classifier is decreased if GAN architectures used in training predate those used during test. 

Previous research \citecustom{Wang_2020_CVPR,gragnaniello2021gan} train classifiers on samples from multiple generators but train each generator on separate data domains or datasets, meaning it is not possible to discern the variation due to the generators themselves. We also note these classifiers are not robust to perturbations and can be fooled with specialized targeted attacks \citecustom{carlini2020evading,goebel2020adversarial}, as is characteristic of classifiers trained using neural networks \citecustom{szegedy2013intriguing}. We consider this out of scope for this work relative to the questions we seek to address.  

\citecustom{yu2019attributing} study GAN attribution, a related problem where the architecture of the source generator for a given sample is inferred. They show that multi-class classification works well to distinguish different GANs, where the learned latent embeddings and weights are used as the image and GAN model fingerprints respectively. \citecustom{marra2019gans} attribute fingerprints to distinguish different GAN model architectures and datasets. These studies show that GAN-generated images contain architecture- and instance-specific artifacts.

There is limited research on the behavior of GAN generators trained to fool such classifiers. \citecustom{chai2020makes} train a specialized patch-based classifier then finetune the GAN generator to fool the classifier, which results in a significant drop in the classifier's accuracy. A second classifier trained using images from the finetuned generator is able to recover in accuracy. \citecustom{zhao2021making} study a related problem of automatically eliminating generator artifacts, training a lightweight CNN generator that adds minimal perturbations to GAN-generated images, allowing them to fool even unseen classifiers. \citecustom{liu2022making} build another such ``trace removal network'' which learns to remove several types of traces left in various types of ``DeepFakes.'' For our study, we are principally interested in the consistency of these artifacts across GAN generators and how the GAN generators adjust themselves when their training loss is modified to include a classifier, to better study the space of artifacts present in GAN-generated images.

To our knowledge, there has not been any work evaluating the effect of classifier model architecture and capacity on the learned artifacts, which we have also studied in this paper.
\cutsectionup
\section{Approach}
\cutsectiondown

We study the phenomena outlined in the introduction by creating and measuring the performance of classifiers trained to detect images sampled from \emph{unseen} generators and subsequently training new generators to fool them, in sequential rounds, forming a chain of generators and classifiers. We do this in one of two settings, first with low dimensional images (MNIST), a simplistic DCGAN, and a basic classifier architecture. In the second setting, we use higher dimensional images (FFHQ), and perform experiments using the unmodified StyleGAN2 (SG2) architecture. Seeking to  minimize sources of variance as much as possible, we limit to a single GAN architecture and a fixed dataset in both settings. We also do not use the ``truncation'' trick \citecustom{karras2019style}, a sample-time heuristic commonly used with the SG2 architecture to improve the output visual quality at the expense of diversity (\emph{see} Supplement for more discussion on this). In the SG2 setting, we test three different widely-used classifier architectures: ResNet-50, Inception-v3, and MobileNetV2. These architectures were chosen for their architectural diversity. All classifiers and generators are trained from scratch, without any pre-training. Supplement provides details about the model architectures and training parameters.

\cutsubsubsectionup
\subsection{A note on terminology} \label{sec:terminology}
\cutsubsubsectiondown

Because our procedure involves both GANs and classifiers, there is potential ambiguity in terminology as GANs themselves are trained with a subnetwork designed to distinguish generated images from natural images, which is commonly called the ``discriminator'', ``adversarial network'', or ``critic'', among others. To keep the text clear, we will refer to subnetworks co-trained with a generator which together comprise a GAN as ``\textbf{discriminators}'', denoted $D$. The networks trained on samples from multiple, independently trained generators are referred to as ``\textbf{classifiers}'', $C$. Each sequential round of training a pool of GANs followed by training classifiers is an ``\textbf{iteration}'' (detailed in Sec. \ref{sec:overview_setup}, and Figs. \ref{fig:experiment_setup_1} and \ref{fig:experiment_setup_2}) and is indexed with a superscript. Iterations are distinct from training steps: during a single iteration, GANs are fully trained, then classifiers are fully trained using those GAN generators. Broadly speaking an ``\textbf{artifact}'' is any property of a generated image that distinguishes it from a real image. By ``\textbf{knowledge gaps}'', we are referring to a specific class of artfacts that reliably occur \emph{across} samples from a generator. Since this class of artifacts is the only one studied in this work, we use artifact and knowledge gap interchangeably.

\cutsubsubsectionup
\subsection{Overview of setup and iterations} \label{sec:overview_setup}
\cutsubsubsectiondown
\begin{figure}[h]
	\centering
    \begin{subfigure}{\linewidth}
        \centering
        \includegraphics[width=0.7\linewidth]{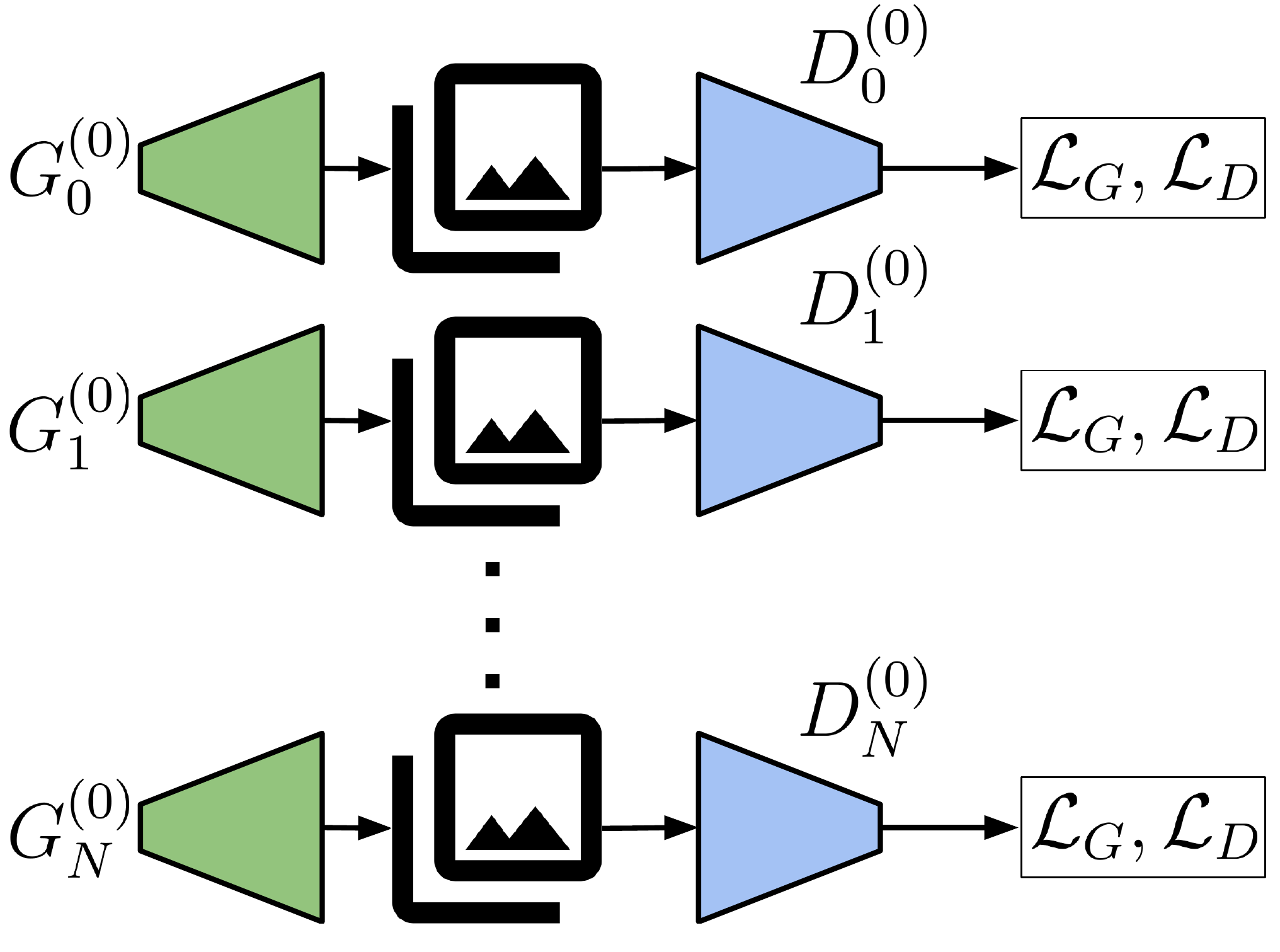}
        \caption{Stage 1 at iteration 0: GAN training with standard loss function} \label{fig:gan_training_zeroth}
    \end{subfigure}%
    \hspace{0.05\textwidth}
    \begin{subfigure}{\linewidth}
    \centering
        \includegraphics[width=0.7\linewidth]{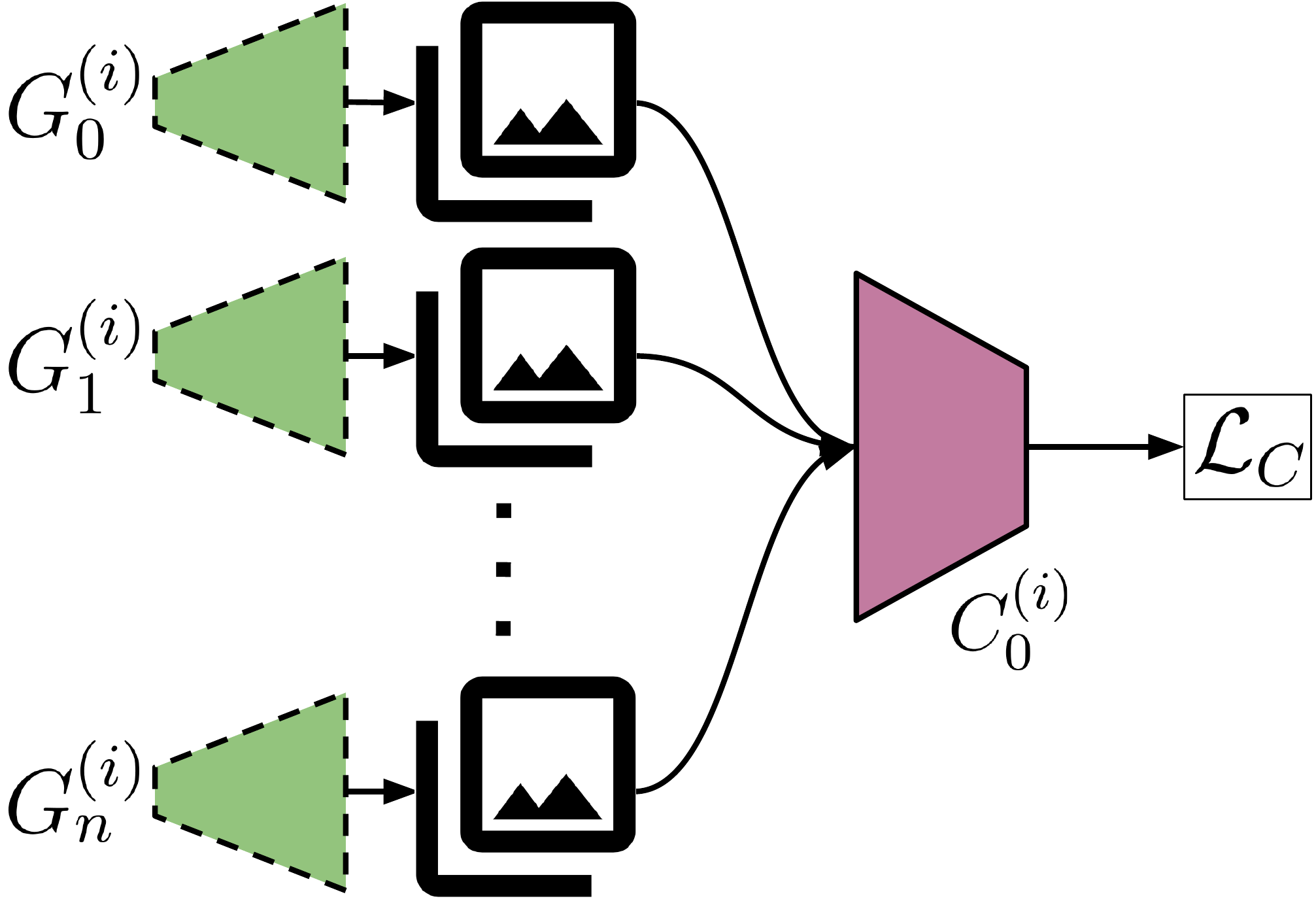}
        \caption{Stage 2 at iteration $i$: Classifier training}\label{fig:classifier_training}
    \end{subfigure}%
	\caption{\textbf{Experimental setup \& training classifiers.} Generators $G$ are \textbf{\textcolor[HTML]{93C47C}{green}}, co-trained discriminators $D$ are \textbf{\textcolor[HTML]{A4C2F4}{blue}} and classifiers $C$ trained using multiple, frozen generators are \textbf{\textcolor[HTML]{C27BA0}{purple}}. Dashed borders indicate that the subnetwork is not being updated during this stage of the iteration. \subref{fig:gan_training_zeroth} Generators trained in iteration 0 are trained in the typical way. \subref{fig:classifier_training} Classifiers are trained in the second stage of all iterations, on samples drawn from subsets of the generators trained in the first stage.}
\label{fig:experiment_setup_1}
\end{figure}

\begin{figure}[h]
	\centering
    \begin{subfigure}{\linewidth}
        \centering
        \includegraphics[width=0.7\linewidth]{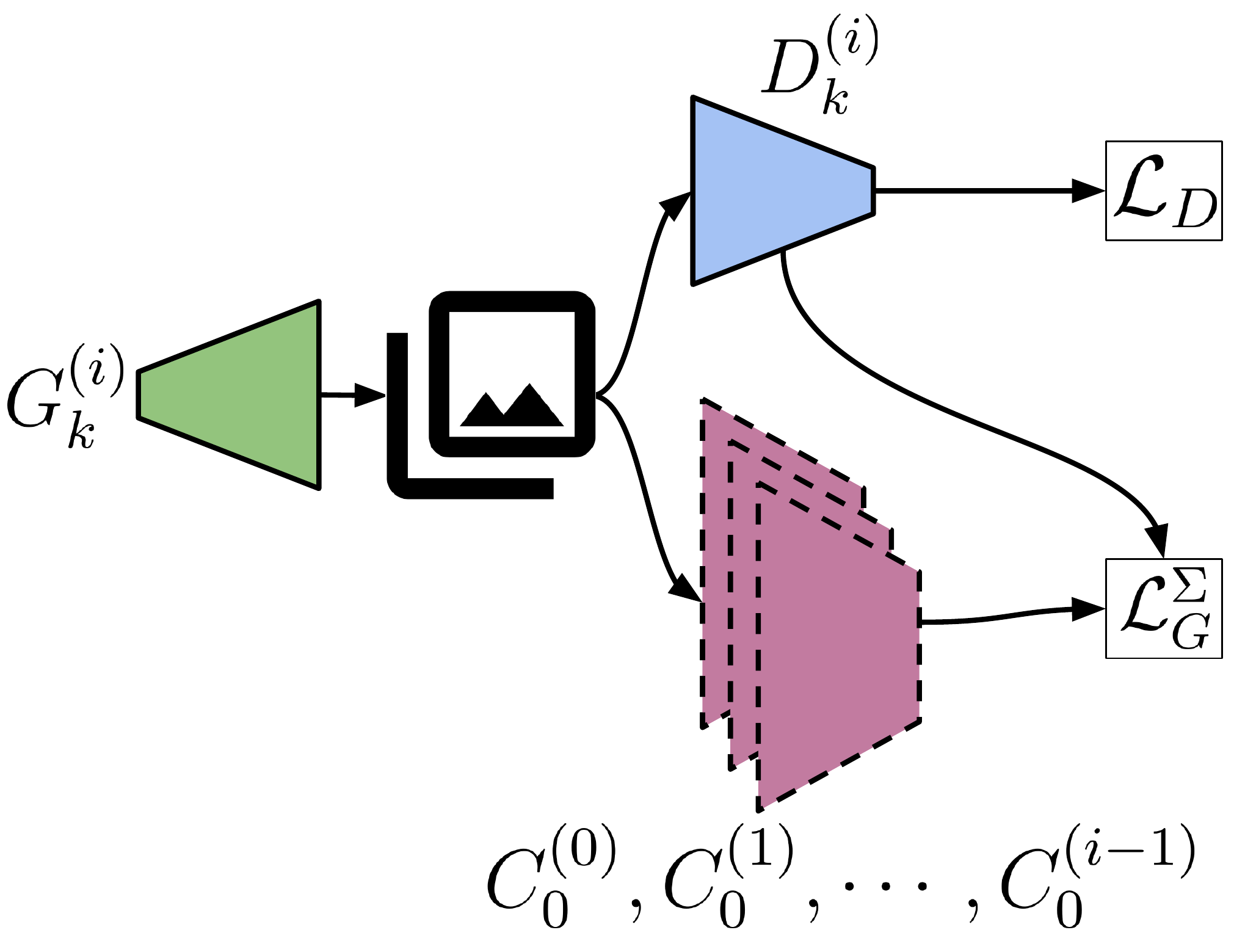}
        \caption{Stage 1 at iteration $i$: GAN training with ``fool-all'' modified loss function}
        \label{fig:gan_training_modified}
    \end{subfigure}%
    \hspace{0.05\textwidth}
    \begin{subfigure}{\linewidth}
        \centering
        \includegraphics[width=0.7\linewidth]{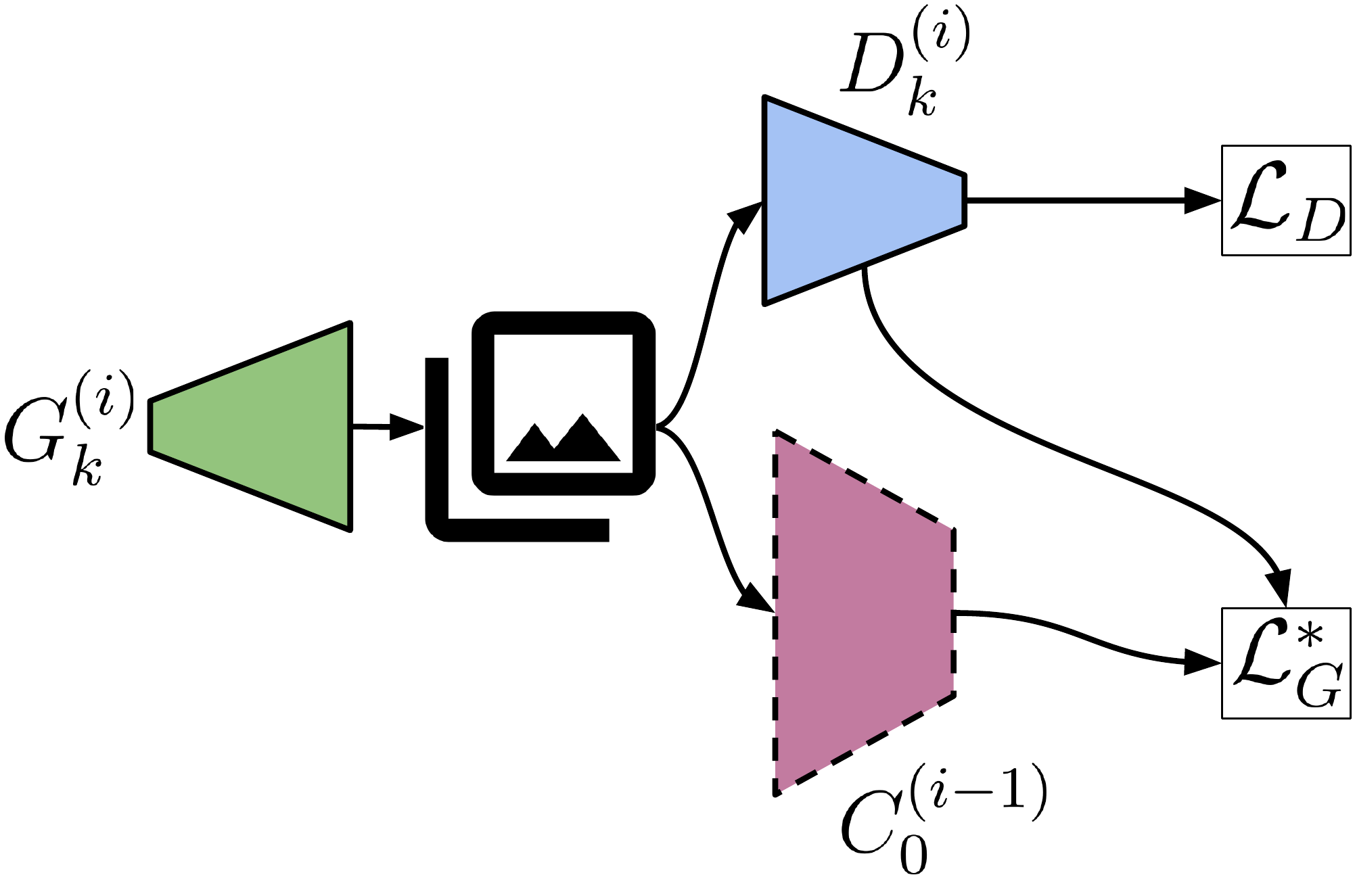}
        \caption{Stage 1 at iteration $i$: GAN training with ``memoryless'' modified loss function} \label{fig:gan_training_modified_memoryless}
    \end{subfigure}%
	\caption{\textbf{GANs trained in higher iterations.} In subsequent iterations ($i > 0$), stage 1 GAN training is modified from the first iteration ($i=0$, \emph{see} \cref{fig:gan_training_zeroth}) such that the generator $G^{(i)}_k$ learns to fool not only its co-trained discriminator $D^{(i)}_k$ but also \subref{fig:gan_training_modified} $i$ classifiers, one from each preceding iteration (\cref{eq:fool_all_gan_loss}) or \subref{fig:gan_training_modified_memoryless} a single classifier from the immediately preceding iteration (\cref{eq:memoryless_gan_loss}). At $i=1$, these two approaches are equivalent.}
\label{fig:experiment_setup_2}
\end{figure}

Our experiments consist of sequential rounds (``iterations''), each with two stages: first, a pool of GAN generators initialized randomly is trained, then classifiers are trained to detect samples from the generators trained in the first stage. In the first stage of the first iteration ($i = 0$), a number of GANs (DCGAN in the first setting, SG2 in the second setting) are trained independently on the training images (MNIST in the first setting, FFHQ in the second setting), as shown in \cref{fig:gan_training_zeroth}. This setup is modified slightly in later iterations (\emph{see} \cref{fig:experiment_setup_2}) as detailed below. Classifier training follows in the second stage (\cref{fig:classifier_training}) as a standard classification task where each classifier is trained on a balanced dataset of real images and images sampled from a subset of generators trained in the first stage. The second stage is the same in every iteration, always sampling images from generators trained in the first stage of the iteration.
The first stage of subsequent iterations ($i > 0$) proceeds like the first stage of the first iteration but with a modified generator loss function: generators are trained to fool not only the discriminator they are co-trained with, but also frozen classifiers from preceding iterations. To do this we modify the ``classical'' GAN generator loss function $\mathbf{\mathcal{L}}$:
\begin{dmath}
\mathbf{\mathcal{L}_{G^{(i)}}} = -\log(D^{(i)}(G^{(i)}(w)))
\label{eq:orig_gan_loss}
\end{dmath}
in one of two ways. In the first, $\mathbf{\mathcal{L}^\Sigma_{G^{(i)}}}$, generators must fool a classifier from \emph{every} preceding iteration: 
\begin{dmath}
\mathbf{\mathcal{L}^\Sigma_{G^{(i)}}} = -[\log(D^{(i)}(G^{(i)}(w))) + \phi \sum_{j=0}^{i-1}\log(C^{(j)}_0(G^{(i)}(w)))]
\label{eq:fool_all_gan_loss}
\end{dmath}
A graphical depiction of a single generator using this loss function is shown in \cref{fig:gan_training_modified}. $\phi$ is a used to weight the relative influence of classifiers. Because a classifier from each previous iteration must be fooled in order to minimize this function, we refer to it as the ``fool-all'' loss function.

The other generator loss function variation, $\mathbf{\mathcal{L}^*_{G^{(i)}}}$, relies purely on a classifier from the iteration immediately preceding the current one, rather than all preceding iterations: 
\begin{align}
\mathbf{\mathcal{L}^*_{G^{(i)}}} = -[\log(D^{(i)}(G^{(i)}(w))) + \phi \log(C^{(i-1)}_0(G^{(i)}(w)))]
\label{eq:memoryless_gan_loss}
\end{align}
This is depicted in \cref{fig:gan_training_modified_memoryless}. Because $\mathbf{\mathcal{L}^*_{G^{(i)}}}$ depends only on the current iteration and the preceding iteration, we refer to this as the ``memoryless'' loss function.

The two modifications result in markedly different training dynamics. Reported results will generally be for the ``fool-all'' $\mathcal{L}^\Sigma$ variation (\cref{fig:gan_training_modified}). When results are based on experiments using the ``memoryless'' variation $\mathcal{L}^*$ (\cref{fig:gan_training_modified_memoryless}), they will be explicitly noted as such. Classifiers are frozen (i.e., their weights are not updated) during the first stage of every iteration.

The classifier subscript $0$, used in Figs. \ref{fig:gan_training_modified} and \ref{fig:gan_training_modified_memoryless} (e.g., $C^{(i-1)}_0$), is purely to distinguish classifiers within the same iteration. In each iteration, multiple classifiers are trained that are initialized randomly and trained independently. When testing a GAN trained to fool the previous iteration's classifiers, classifiers used for training and testing are trained on disjoint subsets of generators, to measure generalization. For example, if $G^{(i)}_k$ is trained to fool $C^{(i-1)}_0$, and is evaluated against $C^{(i-1)}_1$, then $C^{(i-1)}_0$ and $C^{(i-1)}_1$ are trained on disjoint subsets of iteration $i-1$ generators.
\cutsectionup
\section{Results}
\cutsectiondown
We study the interaction between $G^{(0)}$ and $C^{(0)}$ in \cref{sec:diversity_gans}, $C^{(0)}$ and $G^{(1)}$ in \cref{sec:dcgan_fooling_detectors,sec:fooling_detectors}, and compare generators and classifiers of multiple iterations $(C^{(n)}, G^{(n)}, C^{(n+1)}, ...)$ in \cref{sec:training_over_iterations}.

\subsection{Classifiers generalize when sampling training data from multiple generator instances} \label{sec:diversity_gans}
\begin{figure}[h]
    \centering
    \begin{subfigure}{0.5\linewidth}
        \centering
        \includegraphics[width=\linewidth]{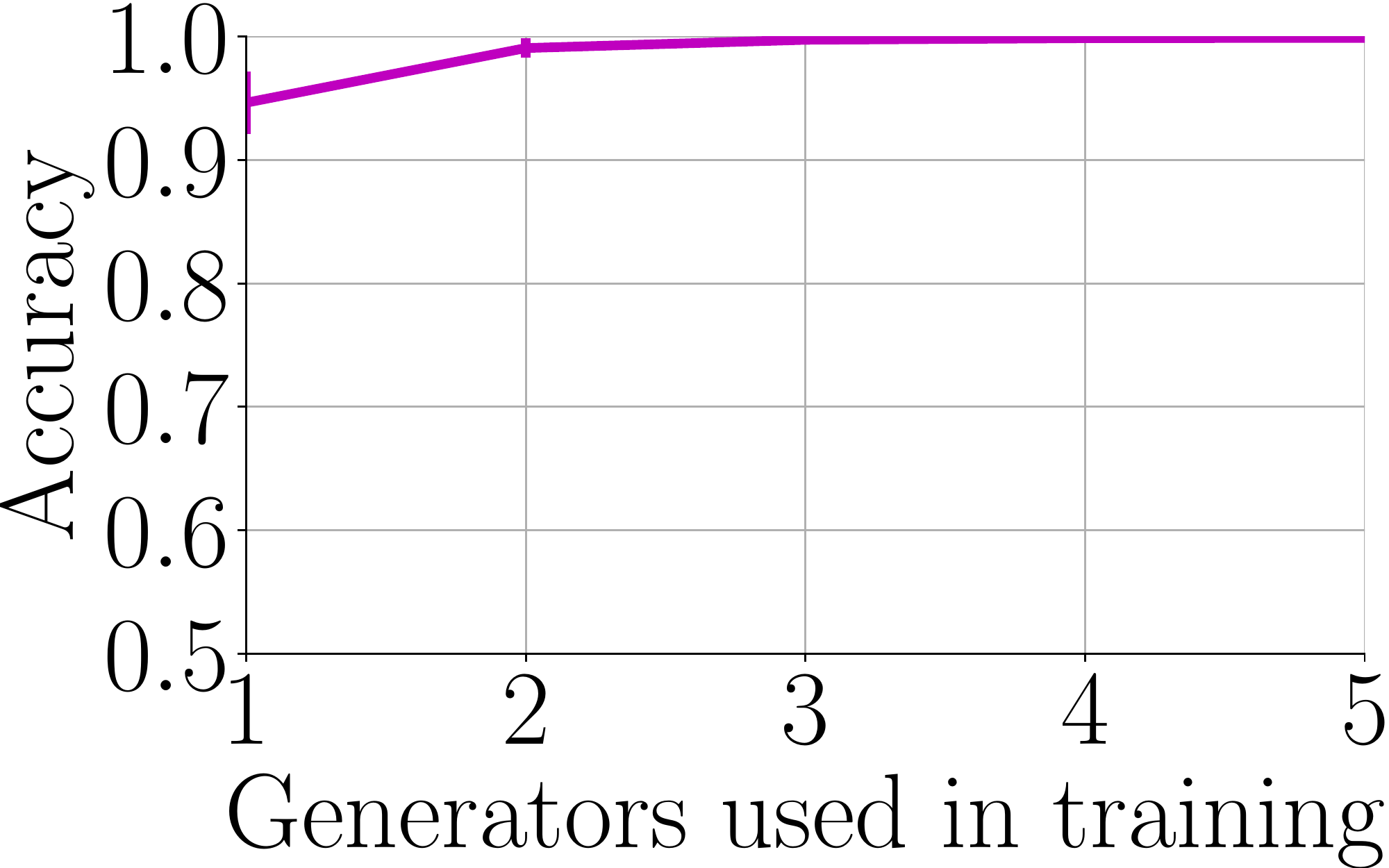}
        \caption{DCGAN classifier}
        \label{fig:generators_diversity_dcgan}
    \end{subfigure}%
    \centering
    \begin{subfigure}{0.5\linewidth}
        \centering
        \includegraphics[width=\linewidth]{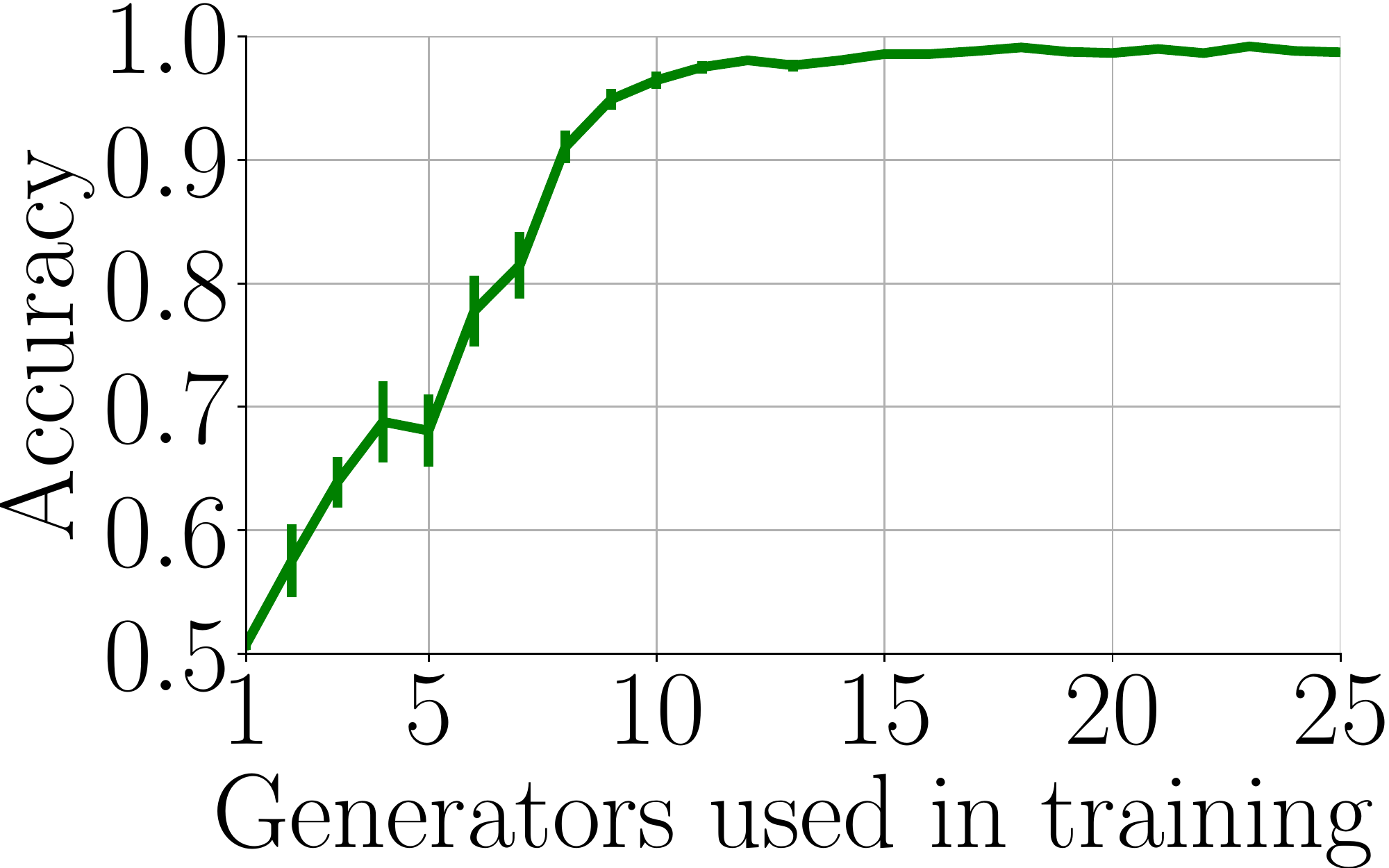}
        \caption{SG2 classifier: ResNet-50}
        \label{fig:generators_diversity_resnet}
    \end{subfigure}%
    \hspace{0.05\textwidth}
    \centering
    \begin{subfigure}{0.5\linewidth}
        \centering
        \includegraphics[width=\linewidth]{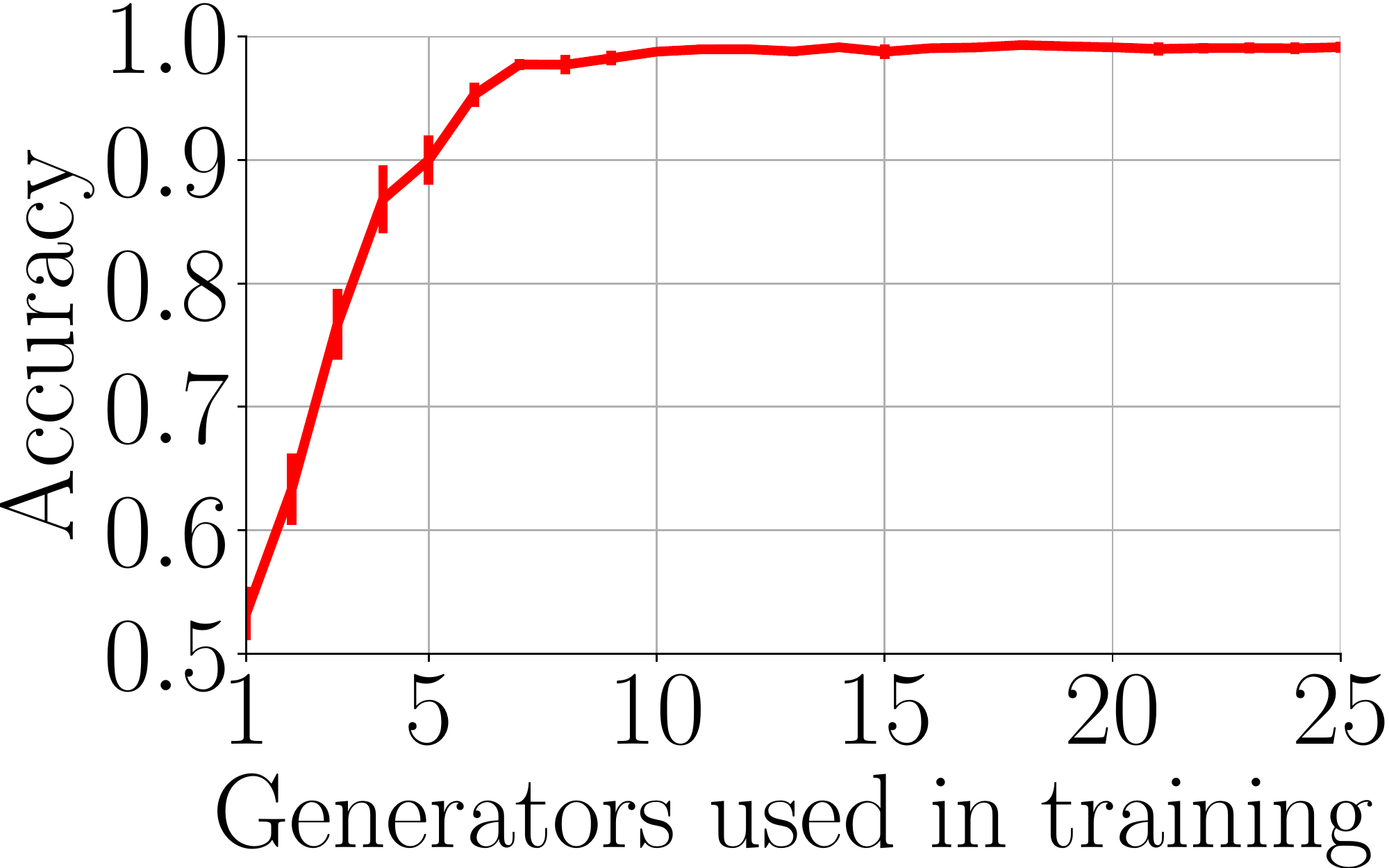}
        \caption{SG2 classifier: Inception-v3}
        \label{fig:generators_diversity_inception}
    \end{subfigure}%
    \centering
    \begin{subfigure}{0.5\linewidth}
        \centering
        \includegraphics[width=\linewidth]{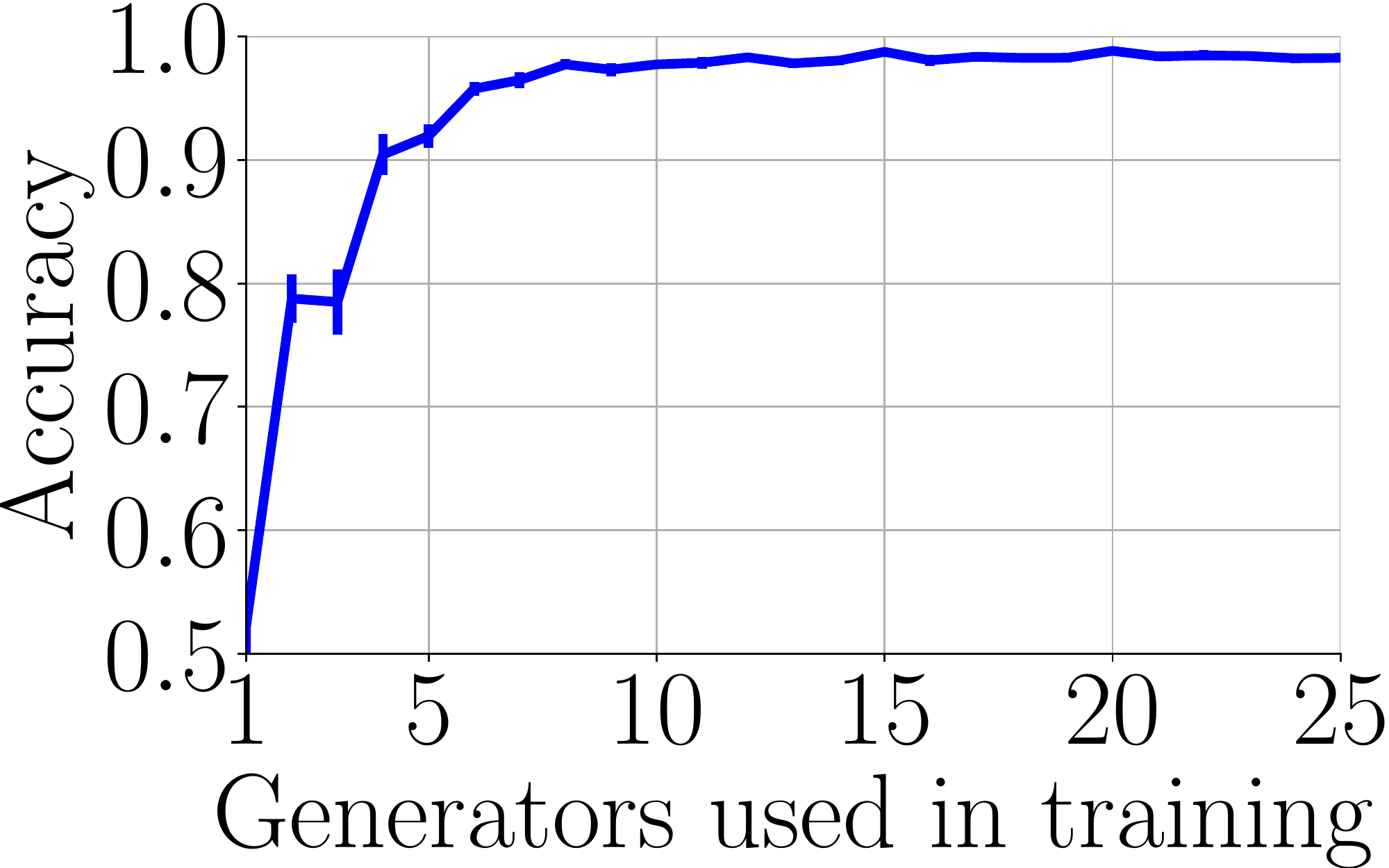}
        \caption{SG2 classifier: MobileNetV2}
        \label{fig:generators_diversity_mobilenet}
    \end{subfigure}%
	\caption{\textbf{Classifier generalization requires samples from several generators.} For all classifier architectures tested, the ability to distinguish between images sampled from unseen generators and real images (y-axis) depends on the number of generators used to produce training data (x-axis). This effect is present in the DCGAN setting \subref{fig:generators_diversity_dcgan} and is amplified in the SG2 setting \subref{fig:generators_diversity_resnet}, \subref{fig:generators_diversity_inception} \& \subref{fig:generators_diversity_mobilenet}. The performance of each classifier is reported as accuracy on a balanced dataset of unseen natural images and samples drawn from a pool of 5 held-out generators in the DCGAN setting, and 25 held-out generators in the SG2 setting; $0.5$ represents random classification.}
\label{fig:multiple_gan_instances}
\end{figure}

Using our pool of generators, we can profile the number of independent generators necessary to train a classifier that can reliably generalize to samples from generators unseen during training. After the first stage of the first iteration, suppose we have a sufficiently large set of trained GAN generators. We split them into two subsets: 
$$S_0 = \{{G^{(0)}_0, ..., G^{(0)}_n}\} \hspace{8pt} \text{and} \hspace{8pt} S_1 = \{{G^{(0)}_{n+1}, ..., G^{(0)}_N}\}$$
We then train a classifier $C^{(0)}_0$ on samples drawn from generators in $S_0$ and measure its performance by testing it on samples from generators in $S_1$ (while $S_1$ can vary in size, to keep results comparable we fix $S_1$ for a given experiment). By measuring the effect of $n$ on the performance of $C^{(0)}_0$ in this way (\cref{fig:multiple_gan_instances}), we can implicitly measure the distinctiveness of artifacts produced by different generators.

We find that a DCGAN classifier generalizes well using a single generator, and almost perfectly when trained using more than one different generators. However, in the second setting, we find that SG2 generators produce sufficient diversity between generator instances that a classifier requires samples from several generators to produce reliable generalization. With sufficient generators to sample from, however, all classifiers become extremely accurate. By contrast, when the so-called ``truncation trick'' \citecustom{karras2019style} is used to generate samples (\emph{see} Supplement), a single generator is sufficient to achieve nearly-perfect classification accuracy on unseen generators. Based on this finding, we use 3 generators in the DCGAN setting and 15 generators in the SG2 setting when training classifiers for the rest of our experiments.

\subsection{DCGAN generators fail to fool classifiers in a generalizable way}\label{sec:dcgan_fooling_detectors}
\begin{figure}[h]
    \centering
    \includegraphics[width=0.9\linewidth]{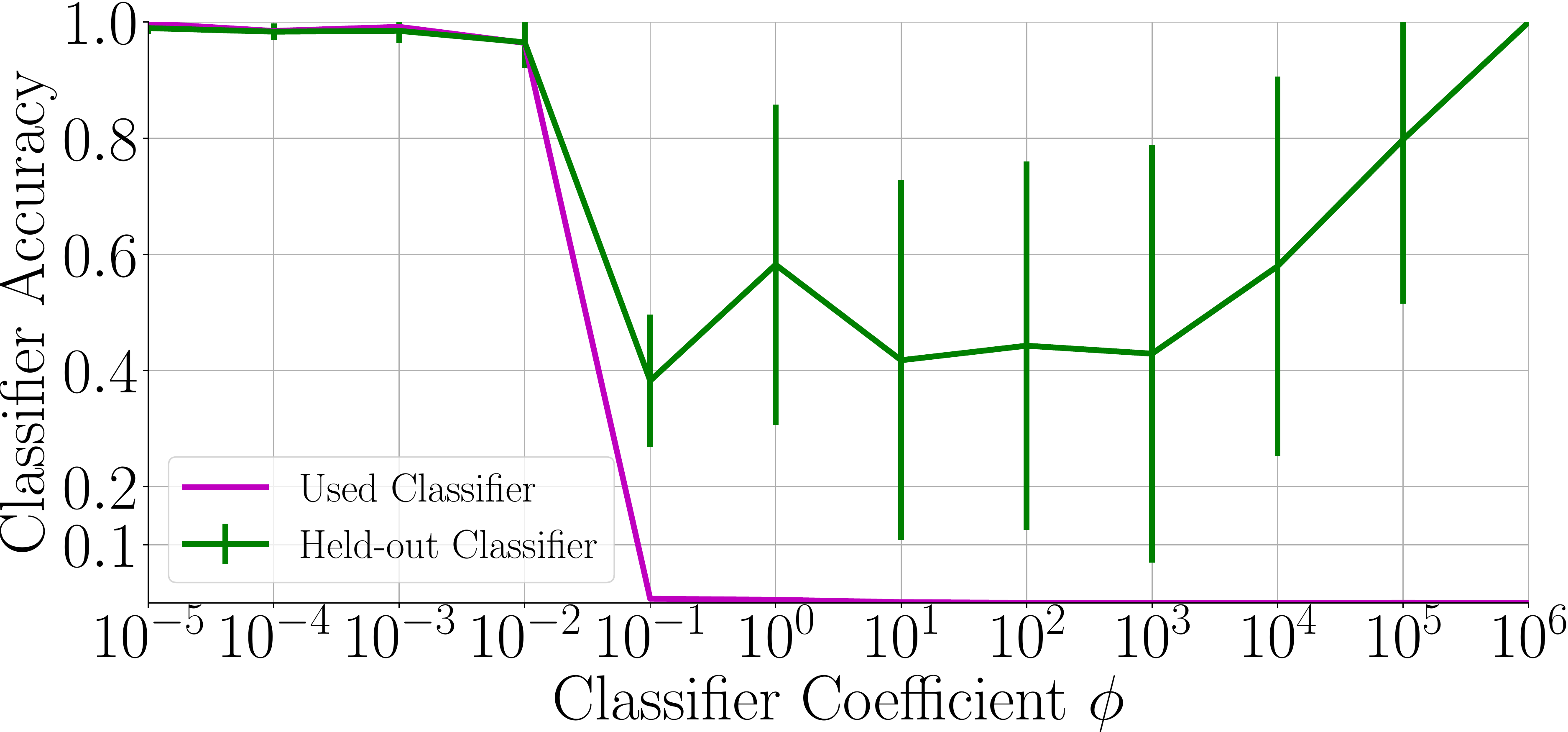}
	\caption{\textbf{DCGAN fails to fool held-out classifiers}. Here, the accuracy is reported on GAN-sampled images \emph{only}, where an accuracy of $0$ represents the GAN completely fooling the classifier. We report the mean accuracy of 5 held-out classifiers.}
\label{fig:dcgan_fooling_accuracy}
\end{figure}

When using \cref{eq:fool_all_gan_loss} to train a DCGAN of the second iteration $i=1$, we find that, surprisingly, the GAN struggles to fool a held-out classifier. This effect is shown in \cref{fig:dcgan_fooling_accuracy} where the DCGAN learns to fool the classifier included in \cref{eq:fool_all_gan_loss} at higher values of $\phi$, but it fails to fool a held-out classifier of iteration $i=0$. As we increase the value of $\phi$ to large values, we see the output quality degrades, as shown in \cref{table:dcgan_output_quality}. Note that because we experiment with higher values of $\phi$ in this section, we normalize the coefficients as:
\begin{align}
\mathbf{\mathcal{L}^*_G} = -[\frac{1}{1+\phi}\log(D) + \frac{\phi}{1+\phi} \log(C)]
\label{eq:normalized_gan_loss}
\end{align}

\begin{table}[h]
\vspace{-0.2in}
\centering
\begin{tabular}{cc}
\toprule
$\phi$ & DCGAN Image Samples \\
\cmidrule{1-2}
\vspace{-0.4em}
$10^{-2}$ & \includegraphics[align=c,scale=0.5]{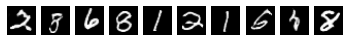} \\
\vspace{-0.4em}
$10^{-1}$ & \includegraphics[align=c,scale=0.5]{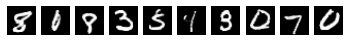} \\
\vspace{-0.4em}
$10^{0}$ & \includegraphics[align=c,scale=0.5]{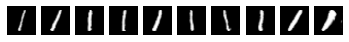} \\
\vspace{-0.4em}
$10^{1}$ & \includegraphics[align=c,scale=0.5]{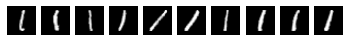} \\
\vspace{-0.4em}
$10^{2}$ & \includegraphics[align=c,scale=0.5]{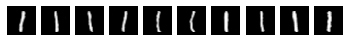} \\
\vspace{-0.4em}
$10^{3}$ & \includegraphics[align=c,scale=0.5]{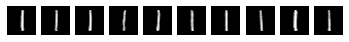} \\
\vspace{-0.4em}
$10^{4}$ & \includegraphics[align=c,scale=0.5]{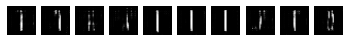} \\
\vspace{-0.4em}
$10^{5}$ & \includegraphics[align=c,scale=0.5]{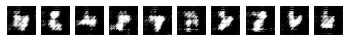} \\
\vspace{-0.1em}
$10^{6}$ & \includegraphics[align=c,scale=0.5]{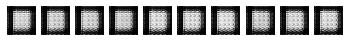} \\
\hline
\end{tabular}
\caption{\textbf{DCGAN output collapses as $\phi$ takes large values.} We show 10 random image samples from each GAN trained with a different value of $\phi$.}
\label{table:dcgan_output_quality}
\end{table}

\subsection{SG2 generators can be trained to fool classifiers in a generalizable way, with caveats}\label{sec:fooling_detectors}
\begin{figure}[h]
    \centering
    \begin{subfigure}{0.5\linewidth}
        \centering
        \includegraphics[width=\linewidth]{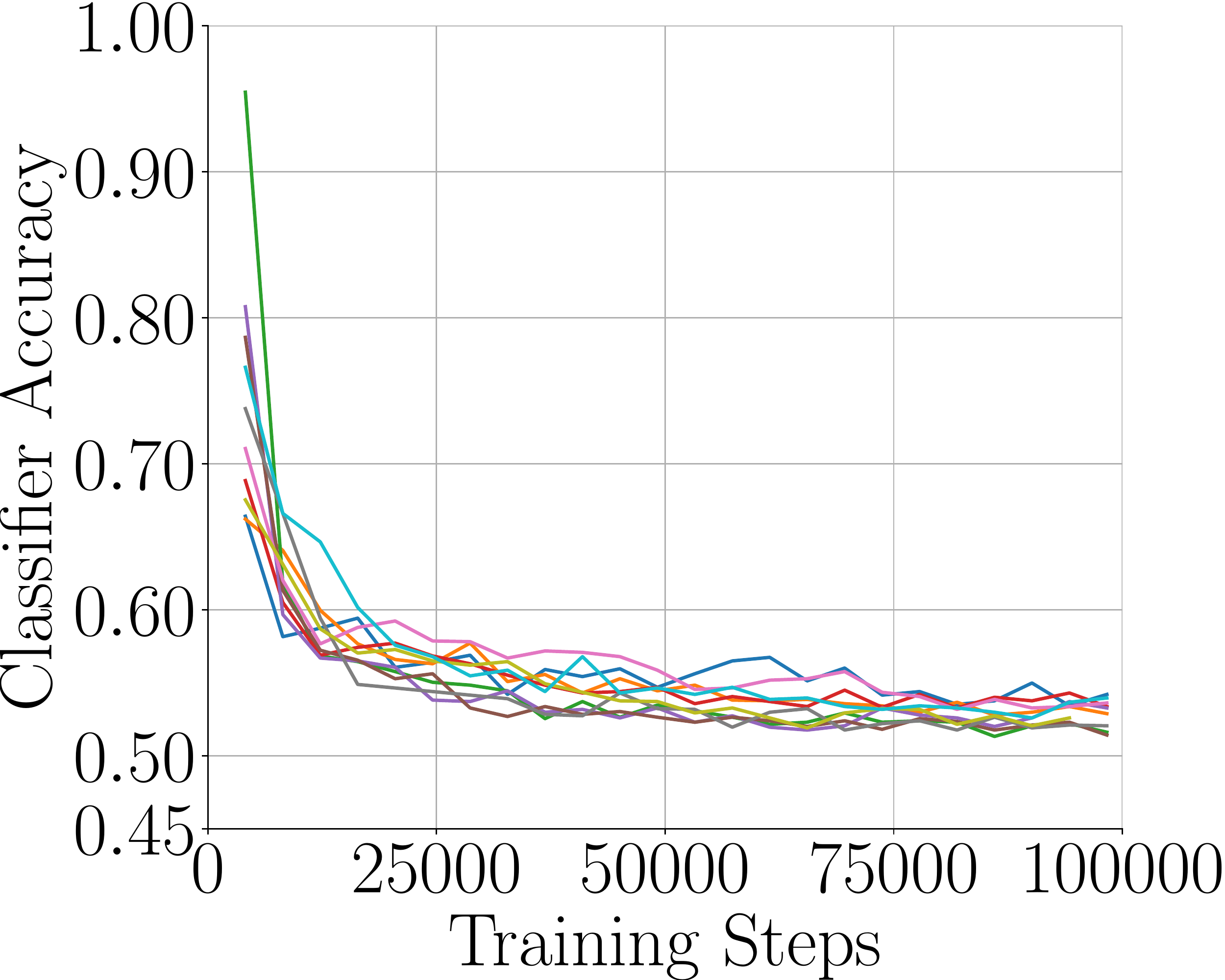}
        \caption{ResNet-50}
    \end{subfigure}%
    \hspace{0.05\textwidth}
    \centering
    \begin{subfigure}{0.5\linewidth}
        \centering
        \includegraphics[width=\linewidth]{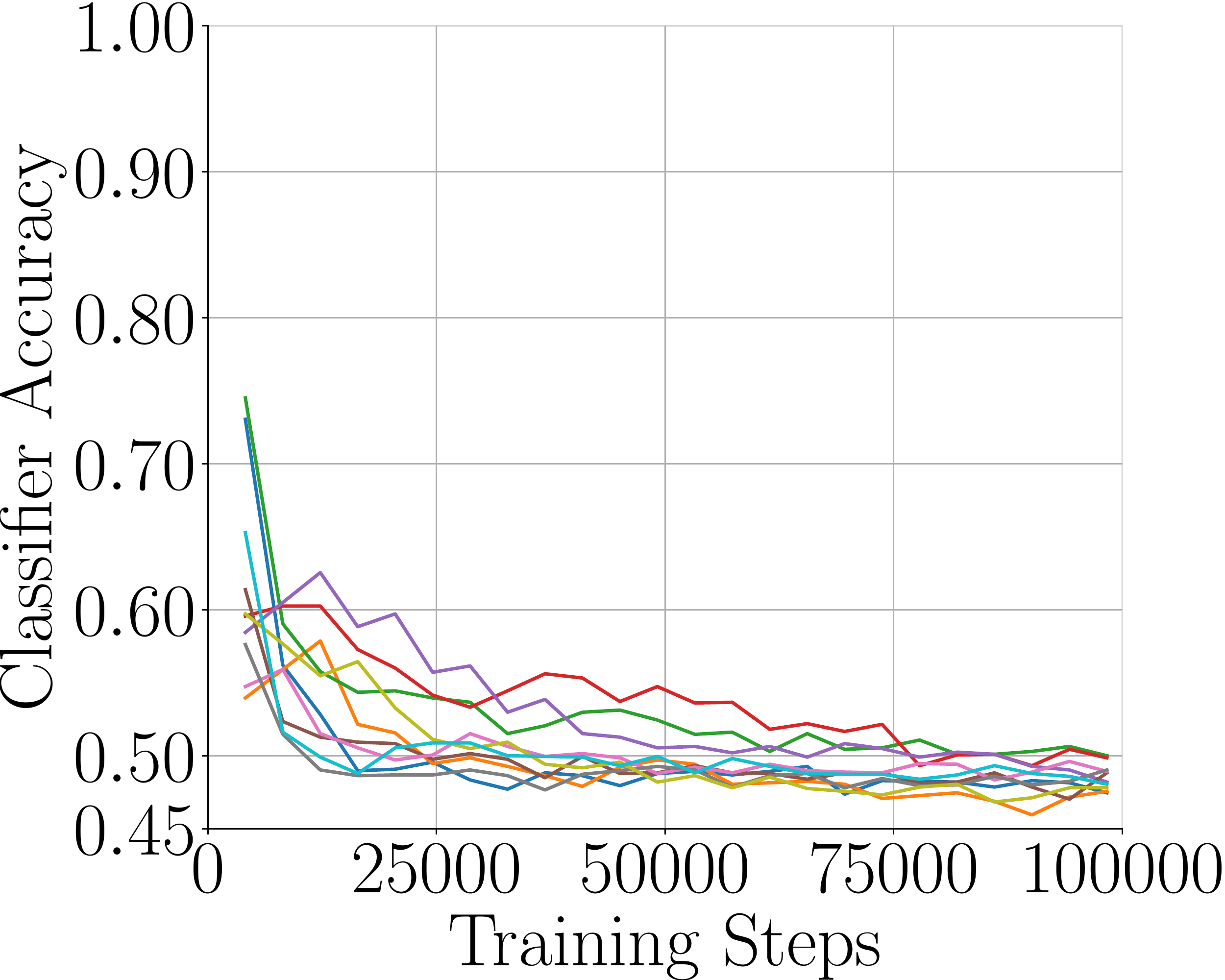}
        \caption{Inception-v3}
    \end{subfigure}%
    \centering
    \begin{subfigure}{0.5\linewidth}
        \centering
        \includegraphics[width=\linewidth]{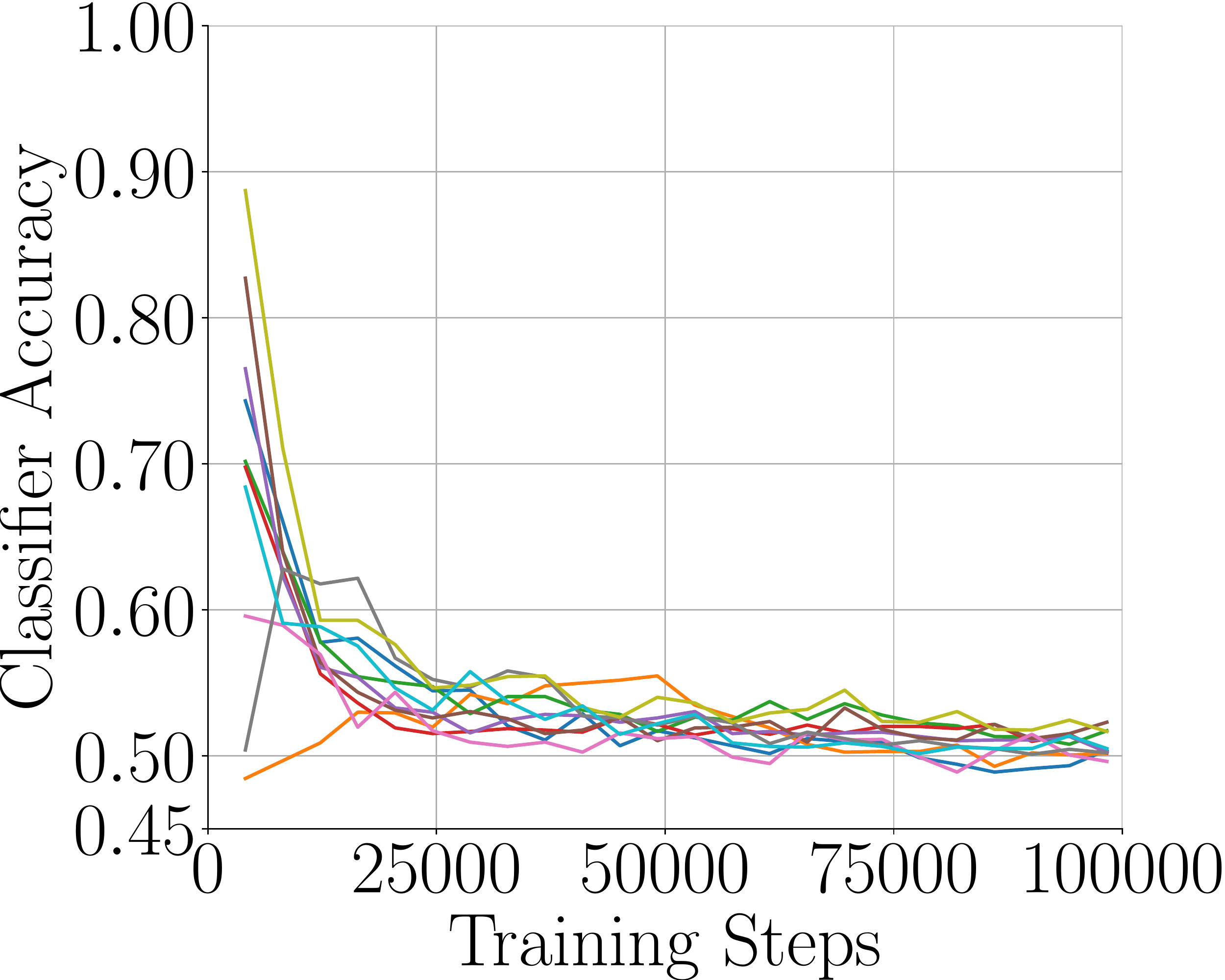}
        \caption{MobileNetV2}
    \end{subfigure}%
	\caption{\textbf{SG2 generators rapidly learn to fool classifiers during training.} Accuracy of first iteration $i=0$ classifiers included in the modified loss with $\phi=0.001$ as a second iteration $i=1$ SG2 generator learns to fool them (at iteration $i=1$, the \cref{eq:fool_all_gan_loss} ``fool-all'' and  \cref{eq:memoryless_gan_loss} ``memoryless'' loss functions are equivalent). We plot 10 runs in each graph, corresponding to 10 different first-iteration classifier instances against 10 different second-iteration generators. Here, classifiers are evaluated against a balanced set of generated and real images; accuracy of $0.5$ is random classification.}
\label{fig:fooling_curve}
\end{figure}

\cutsubsubsectiondown
We observe a different behavior in the SG2 setting than in the DCGAN. In particular, very low values of $\phi$ are sufficient to cause the generator to learn to fool the classifiers. When SG2 generators are trained to fool the classifiers by using the modified loss described in \cref{eq:fool_all_gan_loss} (or \cref{eq:memoryless_gan_loss} where noted), they learn to do so early in their training, as shown in \cref{fig:fooling_curve}.

When using ResNet-50 classifiers, SG2 generators showed a striking ability to generalize: learning to fool one classifier conferred the ability to fool any other ResNet-50 classifier (\emph{see} \cref{table:cross_architecture_eval}). The reliability of this finding, which persists over multiple iterations (\emph{see} \cref{sec:training_over_iterations}), suggests that all ResNet-50 classifier instances $C^{(i)}$ learn strongly overlapping subsets of the artifacts exhibited by the $G^{(i)}$ generators.

However, this is not the case for all classifier architectures tested. The effect was weakened slightly but statistically significantly ($p < 0.01$) in Inception-v3, though generators that fool one Inception-v3 classifier will still fool almost all others. For MobileNetV2, however, the attenuation of the effect was substantial (and statistically significant relative to ResNet-50 and Inception-v3): generators that can fool one MobileNetV2 classifier instance will be able to fool unseen MobileNetV2 classifier instances only half the time. 

We note that our ResNet-50, Inception-v3 and MobileNetV2 architectures had 23.5M, 21.8M, and 2.3M parameters, respectively, and hypothesize that this difference in fooling generalization is an effect of the capacity of the classifier the generator is learning to fool. More concretely, ResNet-50 classifiers, a high capacity model (relatively speaking), each learn the bulk of the artifacts available to them, and thus have high ``knoweldge overlap'' between instances. This overlap accounts for the observed generality of fooling ability on behalf of the generators: it's sufficient to learn to fool one ResNet-50 instance, since all the instances have learned largely the same thing. Conversely, MobileNetV2, a relatively lower capacity model, learns a smaller subset of available artifacts, reducing the probability of overlapping between instances. This reduced overlap means that a generator trained to fool a single MobileNetV2 instance is less likely to fool unseen MobileNetV2 instances when compared to the ResNet-50 case.

Interestingly, the results (\cref{table:cross_architecture_eval}, off-diagonal) imply that the sets of artifacts learned by ResNet-50 and MobileNetV2 are different. If this were not the case, we would expect a MobileNetV2 instance to learn a subset of the artifacts that ResNet-50 learns, due to its lower relative capacity, and hence expect GAN generators that fool unseen ResNet-50 classifiers to be able to readily fool unseen MobileNetV2 classifiers. This effect is even more true of Inception-v3: learning to fool Inception-v3 implies fooling unseen Inception-v3 classifiers but not unseen ResNet-50 classifiers and vice versa. Taken together, this suggests that the higher-capacity architectures learn sets of artifacts that are well-conserved within architecture but are largely distinct between architectures.

To quantify the diversity present in MobileNetV2 classifiers, we modify our ``fool-all'' $\mathbf{\mathcal{L}^\Sigma_{G^{(i)}}}$ loss function to accept multiple MobileNetV2 classifiers (each initialized differently and trained independently) from the previous iteration, rather than just one:
\begin{dmath}
\mathbf{\mathcal{L}^{\Sigma\Sigma}_{G^{(1)}}} = -[\log(D^{(1)}(G^{(1)}(w))) + \phi \sum_{j=0}^{1-1}\sum_{k=0}^{t-1}\log(C^{(j)}_k(G^{(1)}(w)))] \\
= -[\log(D^{(1)}(G^{(1)}(w))) + \phi \sum_{k=0}^{t-1}\log(C^{(0)}_k(G^{(1)}(w)))]
\label{eq:multiMobileNet_loss}
\end{dmath}

\begin{table}[h]
\centering
\begin{tabular}{r@{\hskip 0.1in}c@{\hskip 0.1in}c@{\hskip 0.1in}c@{\hskip 0.05in}c}
\toprule
 & \multicolumn{3}{c}{GAN trained to fool...} \\
 \cmidrule(r){2-4}
Classifier & ResNet-50 & Inception-v3 & MobileNetV2 \\
\midrule
ResNet-50 & \textbf{0.05$\pm$0.02} & 0.74$\pm$0.12 & 0.5$\pm$0.39 \\
Inception-v3 & 0.51$\pm$0.25 & \textbf{0.16$\pm$0.26} & 0.53$\pm$0.34 \\
MobileNetV2 & 0.31$\pm$0.17 & 0.36$\pm$0.16 & 0.41$\pm$0.38 \\
\bottomrule
\end{tabular}

\caption{\textbf{Generalization of SG2 generator classifier-fooling ability varies widely with classifier architecture.} Each entry is the accuracy (mean$\pm$std) of classifiers (row) on generated images \emph{only}, sampled from SG2 generators trained to fool other classifiers (column). Each entry is based on the performance of ten (unseen) classifiers measured against each of ten SG2 generators, where each generator is trained to fool a distinct classifier instance. ``Fooling'' is considered to occur when accuracy $\leq0.20$ (an arbitrary threshold chosen for visualization purposes), and is highlighted in \textbf{bold}. Note that, in all cases, classifiers are at least $98\%$ accurate on held-out real images.}
\label{table:cross_architecture_eval}
\end{table}
\begin{figure}[h]
    \centering
    \includegraphics[width=0.8\linewidth]{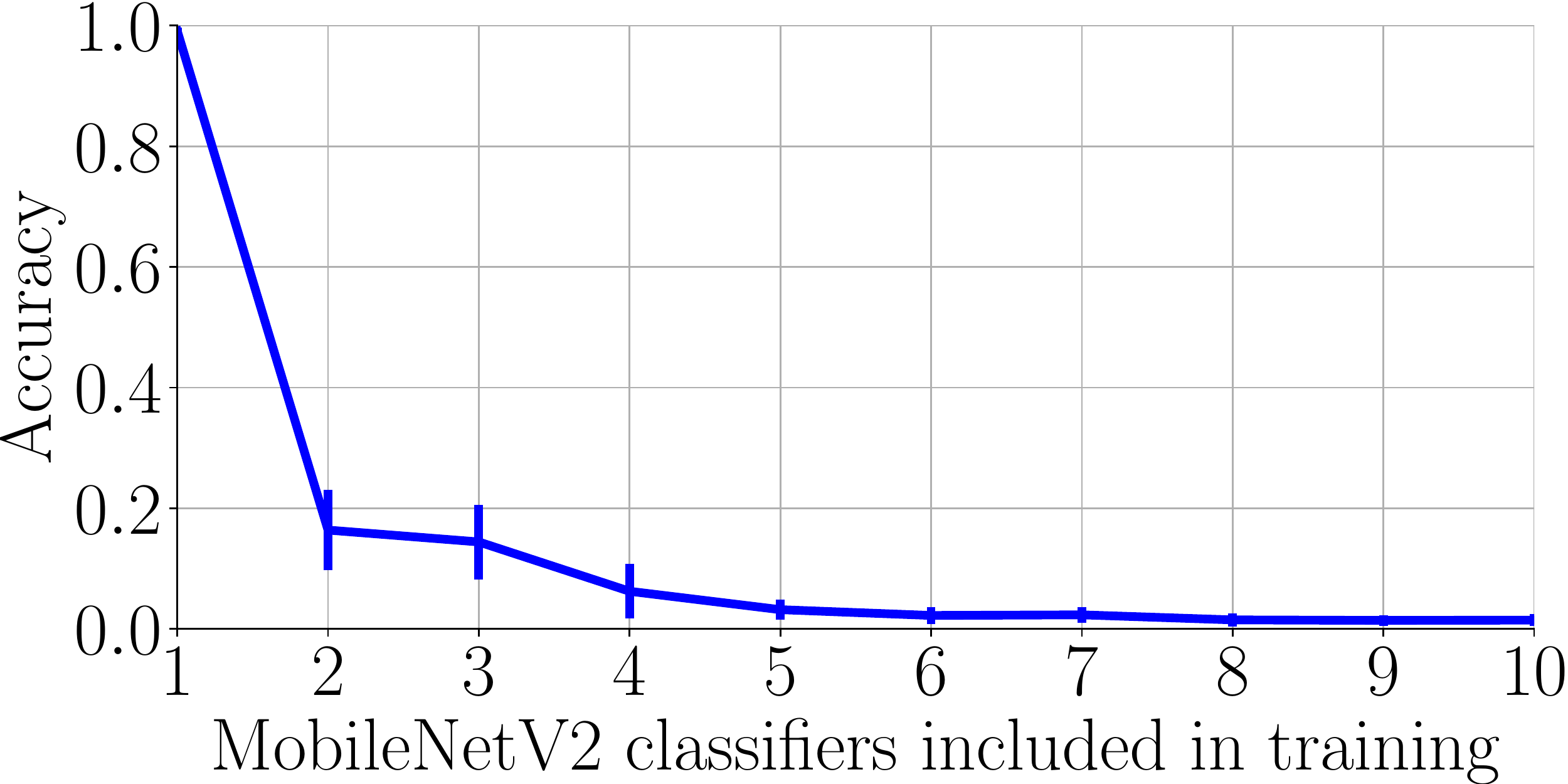}
	\caption{\textbf{Reliably fooling unseen MobileNetV2 classifiers requires learning to fool multiple MobileNetV2 classifiers.} Multiple trained instances of the low-capacity (relatively speaking) MobileNetV2 classifiers are required during generator training to achieve generalizable fooling ability. SG2 generators are trained using the loss function in \cref{eq:multiMobileNet_loss}. We report the mean accuracy of 10 held-out MobileNetV2 classifier instances against generated images from a single trained generator, and we start with fooling a classifier where it completely fails to fool held-out classifiers.}
\label{fig:multiple_mobilenet_instances}
\end{figure}
Because we perform this experiment on iteration $i=1$, there is only one previous iteration (the first, $i=0$), so we may drop the summation-over-previous iteration terms. We add a new summation term to include $t$ classifiers from the previous iteration, rather than a single one. When additional MobileNetV2 classifiers are included, we see the fooling ability of the resulting generators begins to generalize, as shown in \cref{fig:multiple_mobilenet_instances}. 

\begin{table*}[h]
\centering
\begin{tabular}{rcccccccccc}
\toprule
\multirow{2}{*}{MobileNetV2\vspace{7pt}} & \multicolumn{10}{c}{GAN trained to fool MobileNetV2 classifier \#...} \\
\cmidrule(r){2-11}
Classifier & 0 & 1 & 2 & 3 & 4 & 5 & 6 & 7 & 8 & 9\\
\midrule
\cline{2-7}
\rule{0pt}{2.25ex}
0 & \multicolumn{1}{|c}{\textbf{0.04}} & \textbf{0.06} & \textbf{0.05} & \textbf{0.08} & \textbf{0.07} & \multicolumn{1}{c|}{\textbf{0.06}} & 0.98 & 0.98 & 0.47 & 0.95  \\
1 & \multicolumn{1}{|c}{\textbf{0.16}} & \textbf{0.05} & \textbf{0.11} & \textbf{0.10} & \textbf{0.08} & \multicolumn{1}{c|}{\textbf{0.10}} & 0.99 & 0.99 & 0.64 & 0.97  \\
2 & \multicolumn{1}{|c}{\textbf{0.09}} & \textbf{0.08} & \textbf{0.03} & \textbf{0.10} & \textbf{0.08} & \multicolumn{1}{c|}{\textbf{0.08}} & 0.99 & 0.99 & 0.44 & 0.94  \\
3 & \multicolumn{1}{|c}{\textbf{0.09}} & \textbf{0.05} & \textbf{0.05} & \textbf{0.03} & \textbf{0.05} & \multicolumn{1}{c|}{\textbf{0.05}} & 0.99 & 0.99 & 0.53 & 0.96  \\
4 & \multicolumn{1}{|c}{\textbf{0.09}} & \textbf{0.05} & \textbf{0.09} & \textbf{0.06} & \textbf{0.04} & \multicolumn{1}{c|}{\textbf{0.10}} & 0.98 & 0.98 & 0.60 & 0.94  \\
5 & \multicolumn{1}{|c}{0.22} & \textbf{0.15} & \textbf{0.14} & \textbf{0.20} & \textbf{0.20} & \multicolumn{1}{c|}{\textbf{0.05}} & 1.00 & 1.00 & 0.70 & 0.98  \\
\cline{2-9}
\rule{0pt}{2.25ex}
6 & 0.53 & 0.45 & 0.59 & 0.37 & 0.52 & 0.35 & \multicolumn{1}{|c}{\textbf{0.01}} & \multicolumn{1}{c|}{0.21} & 0.37 & \textbf{0.03}  \\
7 & 0.28 & 0.29 & 0.45 & \textbf{0.17} & \textbf{0.17} & 0.21 & \multicolumn{1}{|c}{\textbf{0.02}} & \multicolumn{1}{c|}{\textbf{0.00}} & \textbf{0.12} & \textbf{0.03}  \\
\cline{8-9}
\rule{0pt}{2ex}
8 & \textbf{0.13} & \textbf{0.14} & \textbf{0.06} & \textbf{0.11} & \textbf{0.17} & \textbf{0.09} & 0.92 & 0.91 & \textbf{0.03} & 0.74  \\
9 & 0.44 & 0.42 & 0.37 & 0.34 & 0.40 & 0.33 & 0.60 & 0.70 & 0.28 & \textbf{0.02} \\
\bottomrule
\end{tabular}

\caption{\textbf{MobileNetV2 classifiers appear to form clusters based on the subset of artifacts learned.} Each entry $(i, j)$ is the accuracy of the classifier $i$ (row) against images sampled from SG2 generators trained to fool the classifier $j$ (column). A value of 0.0 means the classifier was completely fooled while a value of 1.0 means the classifier was never fooled. An accuracy $\leq0.20$ (an arbitrary threshold) is considered ``fooled'' and is highlighted in \textbf{bold}. Note that, in all cases, classifiers are at least $98\%$ accurate on held-out real images.}
\label{table:cross_MobileNet_eval}
\end{table*}

Consistent with our classifier-capacity hypothesis, multiple MobileNetV2 classifiers are required to achieve generalization because each instance may only learn a subset of the artifacts available to it, and so do not overlap as often as higher-capacity architectures. To measure the between-instance overlap of MobileNetV2, we conducted a pairwise comparisons over 10 independent MobileNetV2 classifiers. To this end, we trained 10 generators to fool one of 10 trained MobileNetV2 classifiers, then we tested that generator's fooling ability against the other MobileNetV2 instances (as well as the one they were trained to fool). The results (\cref{table:cross_MobileNet_eval}) are striking: rather than randomly sampling artifacts to learn, which would result in fairly uniform off-diagonal values in the table, we see clear clusters emerging that are ``mutually-fooling'': a generator trained to fool one will fool the rest, which we take as implying the classifiers within a cluster learned a shared subset of generator artifacts. For instance, classifier instances 6 and 7 are (almost) mutually fooling, as are classifiers 0 through 5; however, a generator trained to fool 6 or 7 is totally unable to fool classifiers 0 through 5.  In the table, the two clusters mentioned are highlighted using boxes. Perhaps more surprising, the table lacks diagonal symmetry: suggesting that some classifiers learn partial subsets of the artifacts learned by others.

\cutsubsubsectionup
\subsection{SG2 generators of subsequent iterations do not change in image quality or learning behavior}\label{sec:training_over_iterations}
\cutsubsubsectiondown
\cref{sec:diversity_gans,sec:dcgan_fooling_detectors,sec:fooling_detectors} are essentially concerned with the first iteration and the first stage of the second iteration, as detailed in Figs. \ref{fig:experiment_setup_1} and \ref{fig:experiment_setup_2}. If we continue conducting iterations in the SG2 setting, new dynamics emerge. 

Regardless of the loss function used (either Eq. \ref{eq:fool_all_gan_loss} or \ref{eq:memoryless_gan_loss}), the training process does not appreciably change for the generators, nor do they require increased training time. Further, this does not result in a drop in the visual quality of the sampled images either, whether measured qualitatively by visual inspection or quantitatively by FID \citecustom{heusel2017gans}, as included in \cref{table:higher_iteration_detectors}. We do not show the images from the FFHQ dataset or image outputs from models trained on the dataset in this work.


\begin{table*}[h]
\centering
\begin{tabular}{rccccc}
\toprule
 & \multicolumn{5}{c}{GAN Instances} \\
\cmidrule{2-6}
Classifier & Iteration 0 & Iteration 1 & Iteration 2 & Iteration 3 & Iteration 4 \\
\midrule
Iteration 0 & \textbf{0.993$\pm$0.002} & 0.028$\pm$0.004 & 0.037$\pm$0.006 & 0.047$\pm$0.011 & 0.053$\pm$0.007 \\
Iteration 1 & 0.000$\pm$0.000 & \textbf{0.996$\pm$0.005} & 0.013$\pm$0.003 & 0.015$\pm$0.002 & 0.008$\pm$0.002 \\
Iteration 2 & 0.001$\pm$0.000 & 0.734$\pm$0.138 & \textbf{0.957$\pm$0.053} & 0.014$\pm$0.003 & 0.029$\pm$0.007 \\
Iteration 3 & 0.004$\pm$0.001 & 0.868$\pm$0.115 & 0.184$\pm$0.133 & \textbf{0.931$\pm$0.069} & 0.018$\pm$0.004 \\
Iteration 4 & 0.007$\pm$0.001 & 0.713$\pm$0.140 & 0.510$\pm$0.141 & 0.068$\pm$0.027 & \textbf{0.858$\pm$0.112} \\
\midrule
Mean FID & 36.98 & 36.62 & 36.67 & 36.39 & 36.83 \\
\bottomrule
\end{tabular}

\caption{\textbf{Classifier performance in sequential iterations.} We train $5$ iterations of SG2 GANs and their classifier counterparts. The SG2 generators of an iteration are trained with a classifier model of each of the previous iterations (using \cref{eq:fool_all_gan_loss}). $i$th iteration classifier models are trained on $15$ trained instances of $i$th iteration SG2 generators. The entry at $(i, j)$ is the accuracy mean$\pm$std of a iteration $i$ classifier evaluated on images generated with a balanced source of $10$ held-out iteration $j$ SG2 generator instances. A value of $0.0$ means the classifier is always fooled; a value of $1.0$ means it is never fooled. Note that, in all cases, classifiers are at least $95\%$ accurate on held-out real images.}
\label{table:higher_iteration_detectors}
\end{table*}

\cref{table:higher_iteration_detectors} details five iterations, where all classifiers are ResNet-50 architecture and generators are trained with the ``fool-all'' loss $\mathbf{\mathcal{L}^\Sigma_{G^{(i)}}}$ (\cref{eq:fool_all_gan_loss}). In each iteration, generators must fool a classifier from all previous iterations, and accordingly, high-capacity classifiers from iteration $i$ are wholly unable to detect generated images from generators of subsequent iterations $k > i$ (upper-right in \cref{table:higher_iteration_detectors}). The converse is not true: higher-iteration classifiers sometimes, but not always, detect lower-iteration generators (lower-left in \cref{table:higher_iteration_detectors}). This phenomenon lacks a readily identifiable pattern but is highly replicable. For instance, across 10 independent trials, iteration 2 classifiers could detect iteration 1 generators about 73\% of the time, while iteration 4 classifiers could detect iteration 3 generators only 7\% of the time. Furthermore, classifiers of subsequent iterations completely fail to detect GANs of iteration 0 (the first column in \cref{table:higher_iteration_detectors}). The fact that this effect is iteration specific suggests underlying complexity.

Along the diagonal, we see high performance: iteration $i$ classifiers are able to learn to identify images synthesized by iteration $i$ generators, unseen by the classifier during training. We do note a potential trend of diminishing classification test accuracy in later iterations, which suggests the generators do begin to de-correlate after a number of iterations. It is not surprising that the generators still produce artifacts even after training against the classifiers. It is surprising, however, that the generalization ability of the classifiers remains high over the course of several iterations. This means that not only are the generators producing new artifacts but these artifacts are shared with the other randomly-initialized and randomly-trained generator instances of the same iteration. In other words, the generators shift consistently and largely in unison across iterations. Taken together, this suggests the existence of an ``artifact preference'' on the part of the generators. If the generation of some set of artifacts is precluded (for instance, by the need to fool classifiers that have learned them), the generators will not only begin to generate new artifacts but \emph{largely the same} set of new artifacts.

The notion that generators produce artifacts according to an orderly preference, where precluding one set of artifacts leads predictably to the generation of a new set of artifacts, is reinforced by our experiments with iterations trained with the ``memoryless'' loss function (\cref{eq:memoryless_gan_loss}). In this regime, generators must fool only a classifier from the previous iteration. Unsurprisingly, SG2 generators trained in this way fool the previous iteration's classifiers ($G^{(i)}$ fools $C^{(i-1)}$), however, they are detectable by the classifier from \emph{two} iterations ago: $G^{(i)}$ is readily detected by classifier $C^{(i-2)}$. This suggests that, fittingly, in the ``memoryless'' training regime generators oscillate between one of two clusters of artifacts depending on the parity of the iteration.

\cutsectionup
\section{Conclusion} \label{sec:conclusion}
\cutsectiondown

GAN-generated images exhibit ``artifacts'' that distinguish them from real images, even if such artifacts are not apparent to the human eye. Many of these artifacts are ``knowledge gaps:'' rather than being sample-specific, they are present in most or all of the samples from a given generator. We study two settings: using DCGAN with the MNIST dataset, and StyleGAN2 with the FFHQ dataset. Our results suggest that, far from being random or just instance-specific, some artifacts are produced in a regular, repeatable way \emph{across} independent generators of the same GAN architecture (and trained on the same dataset), comprising an ``artifact space.'' This is evident from the fact that all classifiers we tested needed samples from a relatively small pool of generators in order to reliably generalize to unseen generator instances.

Between the two settings, and in stark contrast, DCGAN generators trained on MNIST were unable to fool unseen classifiers without compromising output quality. But StyleGAN2 generators trained on FFHQ learned to fool unseen classifiers of the same architecture with high reliability and often from exposure to a single classifier instance. However, this fooling ability does not generalize to classifier instances of other architectures. This suggests that the subset of the available artifacts learned by a classifier is similar within architecture but different between architecture. Further, StyleGAN2 generators' fooling ability generalizes more reliably when the classifier in question is high capacity (ResNet-50, Inception-v3). Hence, high-capacity architectures learn a large proportion of the artifacts available to them, resulting in correlated behavior. However, a classifier like MobileNetV2 only learns a portion of the artifact space available to them, and so multiple trained classifier instances are required when training a generator in order to reliably fool held-out MobileNetV2 instances. This does not imply that MobileNetV2-based classifiers learn an arbitrary subset of the artifact space: pairwise comparisons indicate that they instead tend to fall into clusters that are ``mutually-fooling.''

When iterating the process in the StyleGAN2 setting, we find that the StyleGAN2 generators continue to quickly learn to fool the classifiers. Similarly, the classifiers require samples from only a few of the newly trained generators to learn how to detect unseen generators reliably. This persistence of detector generalization suggesting that the constraint of needing to fool a classifier induces a consistent transformation on the artifact space across generators (rather than, say, inducing a random transformation specific to each generator). Thus, the StyleGAN2 generators of each new iteration produce artifact spaces that are (mostly) distinct compared to previous iteration but are largely the same \emph{within} the iteration. This suggests an induced preference or ordering over artifact spaces, and merits further study. Our results also hint that this process doesn't continue indefinitely: after a sufficient number of iterations, the StyleGAN2 generators may begin to ``decorrelate'' in terms of their artifact space.

Lastly, we discuss the limitations and societal impact of our work. While we expect our findings to be a property of GAN generators broadly, verifying this is left for future work. In line with previous work \citecustom{Wang_2020_CVPR, gragnaniello2021gan}, it would be instructive to investigate the overlap in the ``artifact spaces'' across different GAN architectures, among multiple iterations. From a misinformation mitigation perspective, we hope our findings will motivate further research into detection of GAN-generated images. However, as with all published work on detection, we cannot conclusively say that there is no potential this could benefit bad actors.
\clearpage


{\small
\bibliographystyle{ieee_fullname}
\bibliography{main}
}

\clearpage

\section*{Supplementary Material}

\subsection*{Data}
We use MNIST or FFHQ (depending on the DCGAN or StyleGAN2 setting) as the dataset of natural images in our experiments. They both consist of 70,000 images.
We use $80\%$ of the data for training, and $20\%$ for evaluation, i.e. we use a fixed sample of 56,000 images for training in all the experiments, and use the rest (14,000) for the evaluation of classifiers. Note that this means that we use the same data of natural images for training both GAN and GAN classifiers, across all iterations. MNIST images are used as $28\times28$ grayscale images, and FFHQ images are used as $256\times256$ RGB images in all our experiments, for both training and evaluation. 
\subsection*{GAN training} \label{sec:supplement_gan_training}
In the DCGAN setting, we trained a simplistic DCGAN architecture well suited for the MNIST generation task (unconditional generation of all digits). Specifically, the generator network is modeled as follows: the random noise variable of $100$ dimensions is passed through a fully connected layer of 12544 units, followed by 3 transposed convolution layers of 128, 64 and 32 units, each with a kernel size of 5, before the final transposed convolution unit for image output. All hidden layers use batchnorm and the LeakyReLU activation.

For the second setting, we have used the unmodified StyleGAN2 as the GAN architecture of choice. StyleGAN2 is generally considered a SOTA GAN model, capable of generating high-resolution, diverse, photo-realistic images, especially of human faces. There are several components and techniques used in its training framework that cause the generated images to be of high quality and of greater diversity. We specifically emphasize the additional inputs to the generator network: a latent code being output by a non-linear mapping network, and the random noise inputs, both are fed to the individual layers of the generator network. These techniques help the outputs capture the stochasticity and variance present in the real world.

We use proprietary implementation of StyleGAN2, which replicates the TensorFlow \citecustom{tensorflow2015-whitepaper} implementation available online (https://github.com/NVlabs/stylegan2). We did not tweak any training parameters. For training a single GAN instance, we use 8 NVIDIA Tesla P100 GPUs and each GAN instance roughly required 1 week to train.

For the modified loss functions, we have used $\phi=0.001$ in all experiments in the StyleGAN2 setting. 

We have not used any pre-training for the GANs in the main paper.

\subsection*{Classifier training}
In the DCGAN setting, we use a basic CNN classifier that performed well for this task. The classifier includes two convolutional layers of 32 and 64 units, each with a kernel size of 3. Both layers use ReLU activation and followed by a max pooling of 2 in both dimensions. We train directly on the grayscale images without any compression.

In the StyleGAN2 setting, we have used ResNet-50 (version 1.5), Inception-v3 and MobileNetV2 architectures when training the classifiers. ResNet-50 is a high-performing CNN architecture, particularly for image classification, and has been shown to be effective for our task in previous research \citecustom{Wang_2020_CVPR,gragnaniello2021gan}. One of the distinguishing features of the ResNet-50 architecture is the use of residual connections, that generally enables efficient learning by sharing information between the hidden layers of a deep network. The two other CNN architectures in our study, Inception-v3 (\cite{szegedy2016rethinking}) and MobileNetV2 (\cite{sandler2018mobilenetv2}), are chosen for their differences to ResNet-50. Like ResNet-50, Inception-v3 is also a large CNN architecture but it does not include residual connections and uses a different "module" that is repeated across the layers. We include MobileNetV2 as a relatively lighter capacity architecture, when compared to the other two architectures.

For the StyleGAN2 classifiers, we use the publicly available implementations as part of the TensorFlow library. When training the classifiers, we pass both the natural and GAN-generated images through JPEG encoding. For each classifier, we train using a single Tesla P100 GPU, and the models roughly require 1 day to train. We do not use any pre-training for the classifiers (unlike \citecustom{Wang_2020_CVPR}), to avoid any external influence in our experiments. We let the classifiers train till they reach convergence, and did not need to finetune the parameters for better performance.

We always train the classifiers on a balanced sample of natural and generated images. Therefore, when we train a classifier using a sample of 15 GAN instances, we train with $15\times56000$ generated and an equal number ($15\times56000$) of natural images, and use 0.5 as the classification decision threshold. For this, the natural images are simply repeated 15 times to obtain a balanced training dataset. When evaluating the fooling ability of GAN generators, the held-out test classifiers are trained using 10 held-out GAN instances. 

\subsection*{Training a StyleGAN2 classifier in the presence of the ``truncation trick''}
The ``truncation trick'' is often used with StyleGAN2 (followed from the StyleGAN model) to avoid generating unrealistic images. The approach shrinks the distribution, in order to remove the regions of low density that might be poorly represented by the GAN model. The expression used to shrink the latent distribution is:
\begin{align}
\mathbf{w'} = \mathbf{\bar{w}} + \psi(\mathbf{w}-\mathbf{\bar{w}})
\end{align}
, where $\mathbf{\bar{w}}$ is the the expected value of the mapped latent space. Here, $\psi$ is the coefficient of truncation: $\psi=1$ implies an absence of truncation and $\psi=0$ would correspond to using the (fixed) expected value of the mapped latent space as the latent input for sample generation. Typically \citecustom{karras2019style,karras2020analyzing}, $\psi=0.5$ is effective in practice.

The truncation trick is used with StyleGAN2 if sampling realistic images, when trained with the FFHQ dataset. We note that we also used this trick when we visually compared the image quality. However, the use of this trick effectively shrinks the diversity and brings the samples closer to the ``average face'' that is learned by the model. And therefore, we don't use the truncation trick in our experiments, since our study is directly measuring the extent of diversity present in the GAN models. 

In our experiments, we have also identified that the FID is negatively affected when employing the truncation trick. The FID (lower is better) for a sample of generated images without truncation is $\sim37$, whereas with truncation ($\psi=0.5$) is $\sim81$.

Since the diversity is significantly reduced if sampling images with truncation, we have also identified that we do not require multiple generators when training a classifier to be able to achieve generalization: i.e., training a classifier using truncated samples from just one SG2 generator instance suffices to detect truncated samples from an independent SG2 generator instance. 

Moreover, we note that by training a classifier which can detect the full range of GAN-generated samples, we also achieve a perfect accuracy when detecting samples generated with the truncation trick.

\subsection*{Finetuning StyleGAN2 to transform to the next iteration}
\begin{figure}[h]
    \centering
    \includegraphics[width=0.6\linewidth]{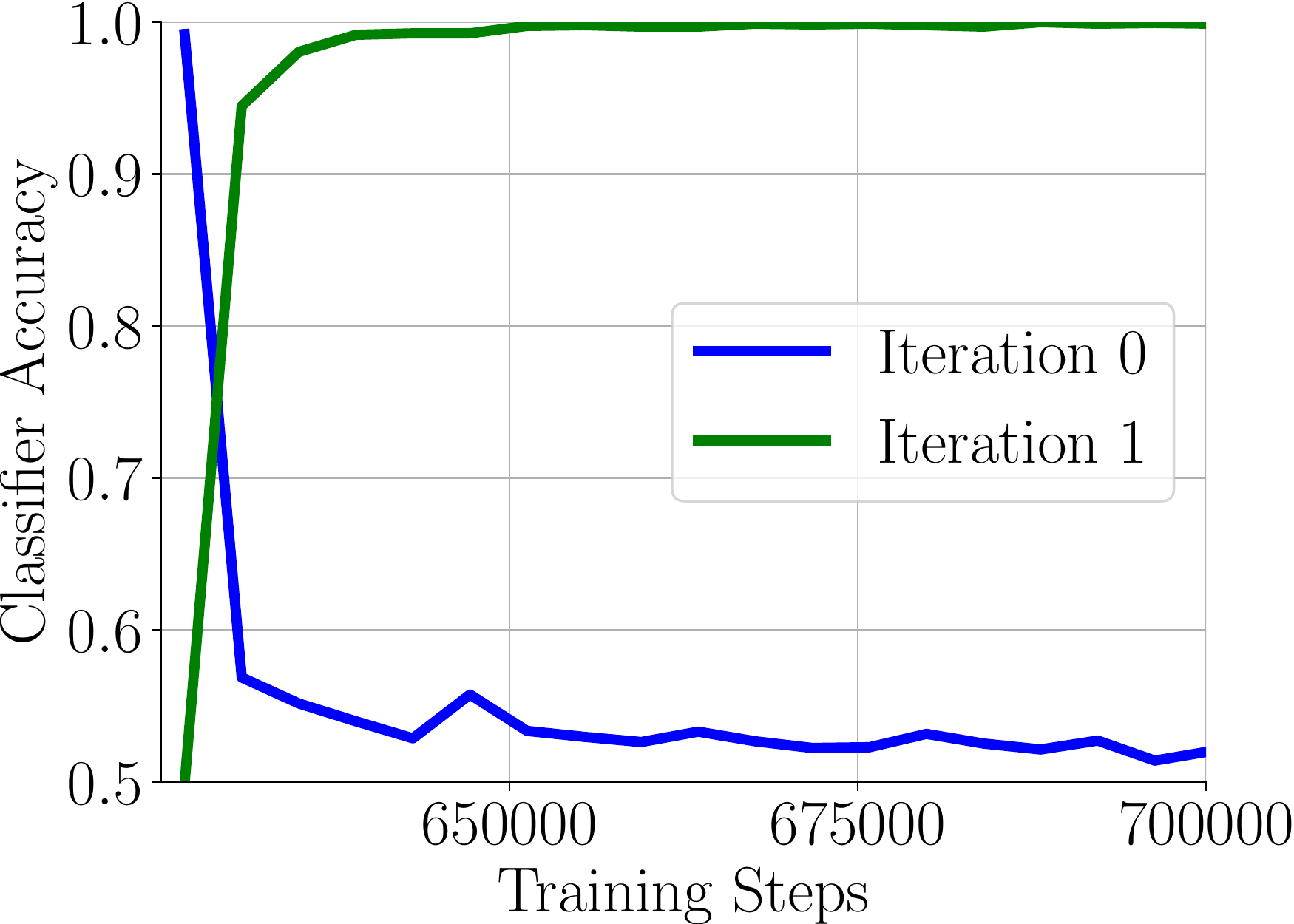}
	\caption{\textbf{Finetuning GAN of iteration $i=\mathbf{0}$.} We finetune a fully trained GAN of iteration 0, using the modified loss in the finetuning steps. As depicted, a held-out classifier of iteration $i=0$ quickly gets fooled, and a held-out classifier of iteration $i=1$ starts to detect the artifacts that were not present before.}
\label{fig:gan_finetuning}
\end{figure}

The experiments mentioned in the paper train GANs from scratch using the modified loss, with random weight initialization. However, we acknowledge that training GANs is an expensive process, where modern GAN models like StyleGAN2 require weeks to train using multiple accelerators. To this point, we have observed that we can ``transform'' a GAN to the succeeding iteration by finetuning a pre-trained GAN, and including a pre-trained detector in the finetuning steps. For instance, using the same modification to the generator loss, finetuning an $i=0$ iteration GAN results in a model that exhibits the same artifacts as what's present in an $i=1$ iteration GAN generator trained from scratch. We arrive at this finding because they are both detected by a held-out $i=1$ iteration detector and they both fool a held-out $i=0$ iteration detector.

\end{document}